\documentclass{new_tlp}

\usepackage[disable]{todonotes}
\usepackage{todonotes}
\def\yu#1{\todo[inline,color=green!40]{#1}}
\usepackage{enumitem}
\usepackage{times}
\usepackage{helvet}
\usepackage{courier}
\usepackage{mathrsfs}
\usepackage[mathscr]{euscript}
\usepackage[all]{xy}
\usepackage{xparse}
\usepackage{verbatim}
\usepackage[all]{xy}

\usepackage{color}
\usepackage{amsmath,amssymb}
\usepackage{latexsym}

\usepackage{xspace}
\usepackage{graphicx}
\usepackage{url}
\newcommand{\ignore}[1]{}

\usepackage{natbib}

\newtheorem{theorem}{Theorem}

\newtheorem{cor}[theorem]{Corollary}
\newtheorem{proposition}[theorem]{Proposition}

\newtheorem{definition}{Definition}
\newtheorem{example}{Example}

\newcommand{\NewASP}{ASP-FO\xspace}

\newcommand{\FO}{FO\xspace}

\newcommand{\Cont}{\mathit{Cont}}
\newcommand{\Aux}{\mathit{Aux}}
\newcommand{\Node}{\mathit{Node}}
\newcommand{\Colored}{\mathit{Colored}}
\newcommand{\cwa}{\mathit{CWA}}
\newcommand{\hd}{\mathit{hd}}
\newcommand{\grnd}{\mathit{grnd}}

\newcommand{\Interview}{\mathit{Interview}}

\newcommand{\Her}[1]{{\mathcal{H}(#1)}}

\newcommand{\struct}[1]{\ensuremath{\langle #1\rangle}\xspace}
\newcommand{\Tr}{\mbox{\bf t}}
\newcommand{\Fa}{\mbox{\bf f}}
\newcommand{\Un}{\mbox{\bf u}}
\newcommand{\Inc}{\mbox{\bf i}}
\newcommand{\I}{\mathfrak{A}}
\newcommand{\J}{\mathfrak{B}}
\newcommand{\Gc}{\mathcal{G}}

\newcommand{\M}{\mathcal{M}}
\newcommand{\dom}[1]{dom{(#1)}}

\newcommand{\In}{\mathit{In}}
\newcommand{\Edge}{\mathit{Edge}}
\newcommand{\Vertex}{\mathit{Node}}
\newcommand{\Nd}{\mathit{Node}}
\newcommand{\Clr}{\mathit{Colour}}
\newcommand{\Cof}{\mathit{ColourOf}}

\newcommand{\vph}{\varphi}
\newcommand{\mim}{\Rightarrow}
\newcommand{\rul}{\leftarrow}
\newcommand{\naf}{\ensuremath{\mathtt{not}\;}}

\newcommand{\x}{\bar{x}}
\newcommand{\xxx}{\bar{x}}
\newcommand{\y}{\bar{y}}
\newcommand{\dd}{\bar{d}}
\newcommand{\ttt}{\bar{t}}
\newcommand{\yyy}{\bar{y}}

\newcommand{\defin}[1]{\ensuremath{\left \{ \begin{array}{l}#1\end{array}
 \right \} }}

\newcommand{\D}{\ensuremath{\mathcal{D}}\xspace}
\newcommand{\extp}[1]{\ensuremath{\mbox{\it Def}({#1})}\xspace}
\newcommand{\exta}{\ensuremath{\mbox{\it Def}}\xspace}

\newcommand{\pars}[1]{\ensuremath{\mbox{\it Par}({#1})}\xspace}
\newcommand{\Voc}{\ensuremath{\Sigma}\xspace}
\newcommand{\ra}{\rightarrow}

\newcommand{\sym}{\tau}

\newcommand{\wh}{\widehat}

\newcommand{\leqt}{\leq_t}

\ignore{

es associated to an ENF sentence
\renewcommand{\equiv}{\Leftrightarrow}

\renewcommand{\vec}[1]{\bar{#1}}

\newcommand{\todo}[1]{$\blacktriangleleft${\bf {#1}}$\blacktriangleright$}

}
\newcommand{\cb}[1]{{\color{blue}#1}}
\newcommand{\cred}[1]{{\color{black}#1}}
\newcommand{\credd}[1]{{\color{green}#1}}

\newcommand{\cL}{\mathcal{L}}
\newcommand{\gen}{{\em generate}\xspace}
\newcommand{\define}{{\em define}\xspace}
\newcommand{\test}{{\em test}\xspace}
\def\beq{\begin{equation}}
\def\eeq#1{\label{#1}\end{equation}}

\newcommand{\Gmodule}{{\sc G}-mo\-dule\xspace}
\newcommand{\Dmodule}{{\sc D}-module\xspace}
\newcommand{\Tmodule}{{\sc T}-module\xspace}
\newcommand{\Gmodules}{{\sc G}-mo\-dules\xspace}
\newcommand{\Dmodules}{{\sc D}-modules\xspace}
\newcommand{\Tmodules}{{\sc T}-modules\xspace}

\newcommand{\coreASP}{core ASP\xspace}

\newcommand{\II}{{\mathcal I}} 

\newcommand{\yulia}[1]{{\bf YULIA} #1 {\bf AILUY}}
\newcommand{\marc}[1]{{\bf MARC} #1 {\bf CRAM}}
\newcommand{\mirek}[1]{{\bf MIREK} #1 {\bf KERIM}}

\NewDocumentCommand{\formulas}{g}{\IfValueTF{#1}{{\mathbb{L}_{#1}}}{{\mathbb{L}}}}
\NewDocumentCommand{\structures}{g}{\IfValueTF{#1}{{\mathbb{S}_{#1}}}{{\mathbb{S}}}}
\NewDocumentCommand{\infsemFO}{mgg}{
\IfValueTF{#2}{\mathscr{FO}_{#1}^{#2}}{\mathscr{FO}_{#1}}
\IfValueTF{#3}{(#3)}{}
}
\NewDocumentCommand{\infsemFOA}{mgg}{
\IfValueTF{#2}{\hat{\mathscr{FO}}_{#1}^{#2}}{\hat{\mathscr{FO}}_{#1}}
\IfValueTF{#3}{(#3)}{}
}
\NewDocumentCommand{\infsem}{mgg}{
\IfValueTF{#2}{\mathscr{I}_{#1}^{#2}}{\mathscr{I}_{#1}}
\IfValueTF{#3}{(#3)}{}
}
\NewDocumentCommand{\infsemGL}{mgg}{
\IfValueTF{#2}{\mathscr{GL}_{#1}^{#2}}{\mathscr{GL}_{#1}}
\IfValueTF{#3}{(#3)}{}
}
\NewDocumentCommand{\infsemT}{mgg}{
\IfValueTF{#2}{\mathscr{OB}_{#1}^{#2}}{\mathscr{OB}_{#1}}
\IfValueTF{#3}{(#3)}{}
}

\title[The informal semantics of Answer Set Programming: A~Tarskian perspective]{The informal semantics of Answer Set Programming: A~Tarskian perspective}

\author[Denecker, Lierler, Truszczynski and Vennekens]
{Marc Denecker\\
Department of Computer Science, KU Leuven, 3001 Leuven, Belgium\\
\email{marc.denecker@cs.kuleuven.be}
\and Yuliya Lierler\\
Department of Computer Science, University of Nebraska at Omaha,
Omaha, NE 68182, USA\\
\email{ylierler@unomaha.edu}
\and Miroslaw Truszczynski\\
Department of Computer Science, University of Kentucky, Lexington, KY 
40506-0633, USA\\
\email{mirek@cs.uky.edu}
\and Joost Vennekens\\
Department of Computer Science, KU Leuven, Campus De Nayer,
2860 Sint-Katelijne-Waver, Belgium\\
\email{joost.vennekens@cs.kuleuven.be}
}

\begin{document}
\label{firstpage}
\maketitle

\begin{abstract}
  In Knowledge Representation, it is crucial that knowledge engineers
  have a good understanding of the formal expressions that they
  write. What formal expressions state intuitively about the domain of discourse 
  is studied in the theory of the informal semantics of a logic. In
  this paper we study the informal semantics of Answer Set Programming.  
  The roots of answer set programming lie in the language of Extended Logic
  Programming, which was introduced initially as an epistemic logic
  for default and autoepistemic reasoning. In 1999, the seminal papers
  on answer set programming proposed to use this logic for a different purpose, namely,
  to model and solve search problems. Currently, the language is used
  primarily in this new role.  However, the original epistemic
  intuitions lose their explanatory relevance in \cred{this new context}. How
  answer set programs are connected to the specifications of problems
  they model is  more easily explained in a classical Tarskian
  semantics, in which models correspond to possible worlds, rather
  than to belief states of an epistemic agent. In this paper, we
  develop a new theory of the informal semantics of answer set programming, which is
  formulated in the Tarskian setting and based on Frege's
  compositionality principle. It differs substantially from the earlier
  epistemic theory of informal semantics, providing a different view
  on the meaning of the connectives in answer set programming and on its relation to other logics,
  in particular classical logic.
\end{abstract}

\section{Introduction}\label{sec:intro}

\begin{quote}{\em I am not here in the happy position of a mineralogist who shows his audience a rock-crystal: I cannot put a {\em thought} in the hands of my readers with the request that they should examine it from all sides. Something in itself not perceptible by sense, the thought is presented to the reader---and I must be content with that---wrapped up in a perceptible linguistic form.}\\
Gottlob Frege, {\it Der Gedanke}
\end{quote}

In knowledge representation, a human expert expresses  informal
propositions about the domain of discourse by a formal
expression in some logic $\cL$. The latter is formulated in
a vocabulary for which the expert has an intended interpretation $\II$
specifying the meaning of the vocabulary symbols in the domain.  It is
essential that the human expert understands which informal
propositions about the problem domain are expressed by formal
expressions of $\cL$. The fundamental task of the {\em informal
semantics} of a logic~$\cL$, sometimes called the {\em declarative
reading} or the \emph{intuitive interpretation} of $\cL$, \textcolor{black}{is
to provide this understanding by explaining formal expressions 
of $\cL$ (or of a fragment of $\cL$) as precise informal propositions
 about the domain of discourse.}

In Frege's terms, an informal proposition is a {\em thought} in the
mind of the expert. It is not a tangible object and is not perceptible
by others. To be observed, studied and used, it must be presented in linguistic form. In this paper, we will therefore assume that the domain expert's intended interpretation $\II$ maps each vocabulary symbol to some natural language statement that represents the imperceptible ``thought'' that this symbol is supposed to represent. Likewise, the informal
semantics of a logic $\cL$ is a mapping that, given such an intended
interpretation $\II$, assigns to a formal expression $\varphi$ of
$\cL$ a natural language reading $\infsem{\II}{}{\varphi}$ that 
captures the meaning of $\varphi$ in $\II$.

We already hint here that natural language is to play a key role in 
any study of informal semantics.
\ignore{In general, it seems unavoidable for natural language to play a key role in any study of informal semantics.} Such use of natural language may be controversial. For instance, \cite{BarwiseCooper81} say: ``To most logicians (like the first author) trained in model-theoretic
semantics, natural language was an anathema, impossibly vague and
incoherent.'' Upon closer inspection, however, the situation is not quite so dire. Indeed, \cite{BarwiseCooper81} go on to say that: ``To us, the revolutionary idea [...] is the claim that natural language is not impossibly incoherent [...], but that large portions of
its semantics can be treated by combining known tools from logic, tools
like functions of finite type, the $\lambda$-calculus, generalized quantifiers,
tense and modal logic, and all the rest.'' In this article, we subscribe to 
this more optimistic view on natural language. While it is certainly possible to create vague, ambiguous or meaningless natural language statements, we believe that a careful use of suitable parts of natural language can avoid such problems. Indeed, much of science and mathematics throughout the centuries has been developed by means of a clear and precise use of natural language. It is this same clarity and precision that we want to achieve in our study of informal semantics. 

The main goal of this paper is to study the informal semantics of answer set programming (ASP) ---  a broadly used
logic-based knowledge representation
formalism~\citep{mar99,nie99,BrewkaET11,journals/ai/GebserKS12}. ASP  has its roots in extended
logic programming~(ELP) proposed by Gelfond and
Lifschitz~(\citeyear{ge1,ge2}). As part of the development of ELP,
Gelfond and Lifschitz presented an informal semantics $\infsemGL{\II}$ for extended
logic programs based on \emph{epistemic} notions of default and
autoepistemic reasoning. According to their proposal, an extended
logic program expresses the knowledge of a rational introspective
agent, where  a stable model (or an answer set) represents a
possible {\em state of belief} of the agent by enumerating all
literals that are believed in that state.
The Gelfond-Lifschitz informal semantics is attuned to applications in
epistemic domains.  However, it is not well aligned with others.  

A decade
after ELP was conceived, researchers realized that, in addition to
modeling applications requiring autoepistemic reasoning, the language
can be used for modeling and solving combinatorial search and
optimization problems \citep{mar99,nie99}.  The term {\em answer set
  programming} was proposed shortly thereafter by
\mbox{\citeauthor{lif99}} (\citeyear{lif99,lif02}) to be synonymous
with the practice of using extended logic programs for this type of
applications.  Since then, ASP has gained much attention and evolved
into a computational knowledge representation paradigm capable of
solving search problems of practical significance
\citep{BrewkaET11}. Particularly influential was the emergence of a
methodology
to streamline the task of 
programming in this paradigm. It consists of arranging program rules in three groups: one to {\em
  generate} the search space, one to {\em define} auxiliary concepts, and
one to {\em test} (impose) constraints.  Lifschitz~(\citeyear{lif02})
coined the term {\em generate-define-test} (GDT) to describe it.
Programs obtained by following the GDT methodology, or \emph{GDT programs},
for short, form the overwhelming majority of programs arising in search
and optimization applications.

However, for GDT programs,
the epistemic informal semantics is inappropriate and ineffective in its
role. To illustrate this point, consider the \emph{graph coloring}
problem. One of the conditions of the problem is the following informal
proposition:
\begin{equation}\label{eq:prop2}
\hbox{\emph{``each node has a color''}}.
\end{equation}
In the language of ELP of 1999, this
condition can be expressed by the rule
\begin{equation}\label{eq:colrule1}
\Aux \leftarrow {\tt not }~\Aux,\ \Node(x),\ {\tt not }~\mathit{\Colored}(x).
\end{equation}
The reading that the $\infsemGL{\II}$ informal semantics provides for 
rule~\eqref{eq:colrule1} is:
\textcolor{black}{``\emph{for every $x$, Aux holds if the agent does not know 
Aux and $x$ is a node and the agent does not know that $x$ has a color.}''} 
There is an obvious mismatch between this sentence and the simple
(objective, non-epistemic) proposition~\eqref{eq:prop2}
that rule~\eqref{eq:colrule1}
 intends to express. In other words, in this
example, the explanatory power of the epistemic informal semantics diminishes.
It fails to provide a direct, explicit link between the formal expression
on the one side, and the property of objects in the domain of discourse
it is intended to represent, on the other.

{Modern ASP dialects typically provide a more elegant notation for writing down constraints, such as:
\begin{equation}\label{eq:colrule2}
\leftarrow  \Node(x),\ {\tt not }~\mathit{\Colored}(x).
\end{equation}
However, in itself this does not address or fix the mismatch. Moreover, as we discuss further on in this paper, it is often surprisingly difficult to extend the Gelfond-Lifschitz epistemic informal semantics to cover the new language constructs of modern ASP dialects.

At the root of the mismatch lies the reflective epistemic agent. 
A key aspect of the original applications for ELP was the presence of such an agent in the domain of discourse; typically it was a
knowledge base that reflects on its own content.} Such an agent is
 absent in the graph coloring problem and in typical
problems that are currently solved using ASP. For example, there are no benchmarks in the series of ASP competitions~\citep{aspcomp1,aspcomp2,aspcomp3,asp4} that mention or require an epistemic
introspective agent.

\textcolor{black}{In this paper, we present a new theory $\infsemT{\II}$ 
of the informal semantics for ASP. 
{We call it \emph{Tarskian} 
because it interprets an answer set of an ASP program in the same way as a model of a first-order logic (FO) theory is interpreted---namely, as an abstraction of a 
{\em possible state of affairs} of the application domain, and not 
epistemically as a state of beliefs of an agent.} We define this theory 
for the pragmatically important class of GDT programs and their 
subexpressions. Our informal semantics explains the formal semantics 
of ASP under the Tarskian view of answer sets. It offers an explanation 
of the meaning of connectives, including ``non-classical'' ones, and it 
satisfies Frege's compositionality principle.
Under the new semantics, the mismatch 
between the information the user encapsulated in the program and the 
intended reading of the program disappears. 
For example,
it maps the constraint~\eqref{eq:colrule2} to the informal proposition~\eqref{eq:prop2}.
It is worth noting that  
the epistemic semantics $\infsemGL{\II}{}$ reflects the fact that
ASP's roots are in the domain of commonsense reasoning. 
{By contrast, the 
informal semantics that we introduce here 
uses the kind of natural language constructs common in mathematical texts.}} 

\ignore{As such, we also specify how ``non-classical'' ASP connectives in GDT programs are to be interpreted to obtain the intended informal semantics.  
The mapping $\infsem{\II}{\cdot}$ is defined for
the expressions in GDT programs using Frege's compositionality
principle.}

A major issue in building an informal semantics for ASP concerns the
structure of programs. Formally, programs are ``flat'' collections of
rules. However, to a human expert, GDT programs have a rich internal
structure. To build our informal semantics $\infsemT{\II}{}$, we develop an
``intermediate'' logic ASP-FO that is directly inspired by the GDT
methodology, in which the (hidden) internal structure of GDT programs is made 
explicit. This structure supports the use of the compositionality
principle when defining the informal semantics for the logic
\NewASP. We show that by exploiting splitting results for ASP \citep{fer09},
programs constructed following the GDT methodology can be
embedded in the logic \NewASP. Thanks to the embedding, our discussion
applies to the fragment of ASP consisting of GDT programs and
establishes an informal semantics for this class of programs.

The paper is organized as follows. \textcolor{black}{We start by reviewing 
theories of informal semantics for two logics: the one of first-order logic 
(Section~\ref{sec:infsem}), which provides guidance for our effort,
and the Gelfond-Lifschitz theory of informal semantics 
for ELP (Section~\ref{sec:infsemGL}), with which we contrast our proposal.}
We then discuss the class of GDT programs, the focus of our work
(Section~\ref{gdt}), and present the logic \NewASP as a way of making 
the internal structure of GDT programs explicit (Section~\ref{sec:newasp}). 
Section~\ref{sec:informal} then presents the main contribution of this paper: the informal semantics $\infsemT{\II}$. Section~\ref{SecASPFOClassical} presents a number of formal results in support of this information semantics. We finish with a discussion of related work (Section~\ref{SecDiscussion}) and some conclusions (Section~\ref{sec:concl}).

\section{The formal and informal semantics of first-order logic}\label{sec:infsem}

\textcolor{black}{ 
In this section, we introduce classical first order logic (FO), with special attention to its informal semantics.  This serves two purposes. First, it is meant as an introduction of the interplay between formal and informal semantics. Much of that will 
be reused for the logic ASP-FO that we define later in this paper. Second,  ASP-FO is a proper superset of FO. Hence, its formal and informal semantics will extend that of FO.}



\paragraph{Formal syntax of FO} 

We assume an infinite supply of \emph{non-logical symbols}: \emph{predicate} 
and 
\emph{function} symbols, each with a non-negative integer 
\emph{arity}. Predicate and function symbols of arity 0 are 
called \emph{propositional} and \emph{object} symbols,
respectively. A \emph{vocabulary} $\Voc$ is a set of \emph{non-logical symbols}.

A term $t$ is an object symbol or a compound expression $f(t_1,\ldots,t_n)$, where $f$ is an $n$-ary function symbol and the $t_i$'s are terms. An atom is an expression $P(t_1,\ldots,t_n)$, where $P$ is an $n$-ary predicate symbol and the $t_i$'s are terms \cred{(in particular, propositional symbols are atoms)}. A formula is then inductively defined as follows:
\begin{itemize}
\item Each atom is a formula;
\item If $t_1, t_2$ are terms, then $t_1 = t_2$ is a formula;
\item If $\varphi_1$ and $\varphi_2$ are formulas, then so are $\varphi_1 \land \varphi_2$, $\varphi_1 \lor \varphi_2$, $\varphi_1 \Rightarrow \varphi_2$ and $\varphi_1 \Leftrightarrow \varphi_2$;
\item If $\varphi$ is a formula, then so are $\lnot \varphi$, $\exists x\ \varphi$ and $\forall x\ \varphi$, where $x$ is an object symbol, here called a {\em variable}. 
\end{itemize}

An occurrence of a symbol $\sym$ in a formula $\varphi$ is \emph{bound} if it is within a subformula of the form $\exists \sym\ \psi$ or $\forall \sym\ \psi$. Otherwise, the occurrence is \emph{free}. In FO, the only symbols that occur bound are object symbols. 

Given a vocabulary $\Voc$, we call a formula $\varphi$ a {\em sentence} over $\Voc$ if all \cred{symbols with free occurrences in $\varphi$}
belong to $\Voc$. The set of sentences over $\Voc$ is denoted $\formulas{\Voc}$.  A \emph{theory}  is a finite set of formulas. A \emph{theory over $\Voc$} is a finite set of sentences over $\Voc$. 

\begin{example} \label{ex:col}
The running application in this section is graph coloring. We choose the vocabulary $\Voc_{col}$ that consists of unary predicates
$\Nd$ and $\Clr$; a binary
predicate $\Edge$; and a unary function $\Cof$ intended to be the mapping from nodes to colors. 
\cred{We define $T_{col}$ to be the following sentence over $\Voc_{col}$:}
\ignore{We introduce the
$\Voc_{col}$-theory consisting of the following sentence
$T_{col}$:}
\begin{gather*}
\forall x\forall y\  (\Edge(x,y) \Rightarrow \lnot \Cof(x) = \Cof(y)).\\
\end{gather*} 
\end{example}

\paragraph{Formal semantics of FO} The basic semantic objects of FO are structures for a given vocabulary.

\begin{definition}
Let $\Sigma$ be a vocabulary. A \emph{$\Voc$-structure} $\I$ consists of (i)
 a non-empty set $\dom{\I}$, called the \emph{domain} of $\I$, and (ii) an 
\emph{interpretation} function $(\cdot)^\I$ that assigns an appropriate 
value $\sym^\I$ to each symbol $\sym$ of $\Voc$:
\begin{itemize}
\item  The value $\sym^\I$ of an $n$-ary function symbol $\sym$ is an
$n$-ary total function over $\dom{\I}$. 
\item The value $\sym^\I$ of an $n$-ary predicate symbol  $\sym$ is an $n$-ary relation over $\dom{\I}$.\footnote{This definition correctly handles the case 
of 0-ary function and predicate symbols. Functions from $\dom{\I}^0$, i.e., 
the empty tuple $()$, 
into
$\dom{\I}$ can be viewed as elements from $\dom{\I}$, yielding a standard 
interpretation of 0-ary function symbols (which are often called constants); 
and each 0-ary predicate symbol is represented by one of exactly two 0-ary 
relations over $\dom{\I}$: {$()$ or $\{()\}$, i.e., false or true}.} 
\end{itemize}
We call $\sym^\I$ the {\em interpretation} or {\em value} of $\sym$ in $\I$.  Given a $\Voc$-structure $\I$, we denote its vocabulary $\Voc$ by $\Voc_\I$.  We say that $\I$ interprets a vocabulary $\Voc'$ if it interprets each symbol in $\Voc'$
(that is, if $\Voc'\subseteq \Voc_\I$). 
\end{definition}
For a given vocabulary $\Voc$, we denote by $\structures{\Voc}$ the class of all $\Voc$-structures. For a $\Voc$-structure $\I$, we define the \emph{projection} of
$\I$ on   $\Voc' \subseteq\Voc$, written $\I|_{\Voc'}$, to be the $\Voc'$-structure
 with the same domain as $\I$ and the interpretation function $(\cdot)^\I$ of $\I$ restricted to $\Voc'$.
We call $\I$ an {\em expansion} of $\I'$ if $\I'=\I|_{\Voc_{\I'}}$.

\begin{example} \label{ex:col:structure}
For the vocabulary $\Voc_{col}$, consider, for instance, the structure $\I_{col}$ defined as follows:
\begin{itemize}
\item $\dom{\I_{col}}= \{n_1,n_2,n_3,c_1,c_2\}$ 
\item  $\Nd^{\I_{col}} = \{n_1,n_2,n_3\}$, 
\item $\Clr^{\I_{col}} = \{c_1,c_2\}$, 
\item $\Edge^{\I_{col}} = \{ (n_1,n_2), (n_2, n_3)\}$ and
\item  $\Cof^{\I_{col}} =\{ n_1\mapsto c_1, n_2\mapsto c_2, n_3 \mapsto c_1, c_1\mapsto c_1, c_2\mapsto c_2\}$.
\end{itemize}
\end{example}
Note that, in FO, a structure must interpret each function symbol by a \emph{total} function on its domain. \cred{Therefore,} while 
we intend $\Cof$ to be a function that maps nodes to colors, it 
 must also map each of the colors to some value. This could be avoided by, for instance,  allowing partial functions, or by using a typed variant of~FO. 
For simplicity, however, we stay with classical definitions of FO.

For a structure $\I$, a 0-ary function symbol $\sym$ 
and a corresponding value $v\in \dom{\I}$,
we denote by $\I[\sym:v]$ the structure identical to $\I$ except that $\sym^{\I[\sym:v]}=v$. 

\begin{definition}
We extend the interpretation function $(\cdot)^\I$ of structure $\I$ to all compound terms over $\Voc_\I$ by the inductive rule\\
--  $f(t_1,\dots,t_n)^\I = f^\I(t_1^\I,\dots,t_n^\I)$. \\
We further extend this function to tuples $\ttt$ of terms over $\Voc_\I$ by defining \\ 
-- $\ttt^{\ \I}= (t_1^{\I},\ldots,t_k^{\I})$, 
where $\ttt=(t_1,\ldots,t_k)$.
\end{definition}
The value/interpretation $t^\I$ is a well-defined object in the domain of $\I$ provided $t$ is a term and $\I$ interprets all function and object symbols in $t$.

Next, we define the truth relation, or the satisfaction relation, between structures and formulas.

\begin{definition}[Satisfaction relation $\I\models 
\varphi$] \label{deftrue}
Let $\varphi$ be an FO formula and $\I$ a  
structure interpreting all free symbols of $\varphi$. We define  $\I\models \varphi$ by 
induction on the structure of $\varphi$:
\begin{itemize}
\item[--] $\I\models P(\ttt)$, where $P$ is a predicate 
symbol, if $\ttt^{\ \I}\in P^\I$;
\item[--]  $\I\models \psi\land\varphi$ if $\I\models 
\psi$ and  $\I\models \varphi$;
\item[--]  $\I\models \psi\lor\varphi$ if $\I\models 
\psi$ or  $\I\models \varphi$ (or both);
\item[--]   $\I\models \neg \psi$ if $\I \not\models \psi$; i.e., if it is not the case that $\I \models \psi$;
\item[--]  $\I\models \psi\Rightarrow\varphi$ if 
$\I \not \models \psi$ or $\I\models \varphi$ (or both);
\item[--]  $\I\models \exists x\ \psi$ if for some 
$d\in\dom{\I}$, $\I[x:d]\models \psi$.
\item[--]  $\I\models \forall x\ \psi$ if for each
$d\in\dom{\I}$, $\I[x:d]\models \psi$.
\item[--]  $\I\models t_1 = t_2$ if ${t_1}^\I = {t_2}^\I$; i.e., ${t_1}, t_2$ have identical interpretations in $\I$. 
\end{itemize}

When $\I\models\varphi$, we say that $\varphi$ is true in $\I$, or that $\I$ satisfies $\varphi$, or that $\I$ is a model of $\varphi$.
\end{definition}
The satisfaction relation is easily extended from formulas to theories.

\begin{definition}
Let $T$ be an FO theory  over $\Voc$. A $\Voc$-structure~$\I$ is a model of $T$ (or satisfies $T$), denoted
$\I\models T$, if $\I\models \varphi$ for each $\varphi\in T$.
\end{definition}

The satisfaction relation induces definitions of several
other fundamental semantic concepts. 

\begin{definition} [Derived semantic relations]
A theory (or formula) $T$ \emph{entails} a formula $\varphi$, denoted $T\models\varphi$, if $\varphi$ is satisfied in every structure $\I$ that interprets all 
symbols with free occurrences in~$T$ and $\varphi$, and satisfies $T$. 
A formula $\varphi$ is \emph{valid} (denoted $\models\varphi$) if it is satisfied in every structure that interprets its free symbols. A formula or theory is {\em satisfiable} if it is satisfied in at least one structure. A formula $\varphi_1$ is {\em logically equivalent} to $\varphi_2$ (denoted $\varphi_1\equiv\varphi_2$)  if for every structure $\I$ that interprets the free symbols of both formulas, $\I\models\varphi_1$ if and only if $\I\models\varphi_2$. 
\end{definition}

\paragraph{The informal semantics of FO}


The theory of informal semantics of FO is a coherent system of interpretations
of its formal syntax and semantics that \textcolor{black}{explains formulas as objective propositions about the application domain, structures as ``states of affairs'' of the application domain, and the satisfaction relation as a truth relation between states of affairs and  propositions. It also induces informal semantics for the derived semantical relations.}  We will denote this informal semantics by $\infsemFO{}$, 
and the three components of which it consists (i.e., the interpretation of 
formulas, the interpretation of structures, and the interpretation of 
semantic relations such as satisfaction) by $\infsemFO{}{\formulas}$, 
$\infsemFO{}{\structures}$ and  $\infsemFO{}{\models}$, respectively.

The informal semantics of a formula $\varphi$ is the information that 
is represented
by $\varphi$ about the problem domain. It is essentially a
\emph{thought}. Following the quote by Frege at the beginning of this
article, we make these thoughts tangible by giving them a linguistic form. 
In other words, the first component of the theory of
informal semantics of FO consists of a mapping of FO
formulas to natural language statements. 

The informal semantics of a formula $\varphi$ depends on a parameter ---
the meaning that we give to the symbols of vocabulary $\Voc$ of $\varphi$ 
in the application domain. 
This is captured by the {\em intended interpretation} $\II$ of the vocabulary $\Voc$. To state the informal semantics of a formula over $\Voc$ in linguistic form, we specify $\II$ as an assignment of natural language expressions  to the symbols of $\Voc$.
For an $n$-ary function $f/n$, $\II(f/n)$ (or $\II(f)$, if the arity is clear 
or immaterial) is a parameterized noun phrase that specifies the value of the function 
in the application domain in terms of its $n$ arguments.  Similarly, for an $n$-ary predicate $p/n$, $\II(p/n)$ is a parametrized sentence describing the relation between $n$ arguments of $p$. In either case, the $i$th argument is denoted as $\#_i$.

\textcolor{black}{
\begin{example}
In the running example, the  intended interpretation~$\II_{col}$ of 
the vocabulary~$\Voc_{col}$ can be expressed in linguistic form as 
parameterized declarative sentence and parameterized noun phrases:
\begin{itemize}
\item $\II_{col}(\Nd/1) =$``$\#_1$ is a node'';   
\item $\II_{col}(\Edge/2) =$``there is an edge from $\#_1$ to $\#_2$'';
\item $\II_{col}(\Cof/1) =$``the color of $\#_1$'';
\end{itemize}
\end{example}
}

\ignore{
Recall, that set $\formulas{\Voc}$ denotes the set of all $\Voc$-formulas. By 
$\structures{\Voc}$, we denote the set of all $\Voc$-structures. The three components
$(\formulas{\Voc}, \structures{\Voc},\models)$  form the
logic FO. A theory of informal semantics explains the meaning of each of these
three components, i.e., it explains the meaning of the formal
syntactical objects, of the formal semantical objects, and of the
satisfaction relation. (Typically, the first of these three points is
the one which requires the most attention; nevertheless, the other two are also necessary.)
We now address each of these three points for FO.
}

Given an intended interpretation $\II$ for a vocabulary $\Voc$, the informal semantics $\infsemFO{\II}{\formulas}$ of FO terms and formulas over $\Voc$ is now the inductively defined mapping from formal expressions to natural language expressions specified in Table~\ref{fig:FOinf:form}. 

\begin{table}[ht]
\caption{The informal semantics of FO formulas.\label{fig:FOinf:form}}
\begin{center}
\begin{tabular}{ccp{7cm}}
\hline\hline
\hspace{-0.3in}$\Phi$ & & \hspace{2.5cm}$\infsemFO{\II}{\formulas}{\Phi}$\rule{0pt}{0.3cm}\\
\hline\hline
\hspace{-0.3in}$x$ & & x \emph{(where $x$ is a variable)}\rule{0pt}{0.2cm}\vspace{0.0cm}\\
\hline
\hspace{-0.3in}$f(t_1,\ldots,t_n)$ & \phantom{aaaaa}&\begin{minipage}{8cm}$\II(f)\langle \infsemFO{\II}{\formulas}{t_1},\ldots,\infsemFO{\II}{\formulas}{t_n}\rangle$\\\emph{(i.e., the noun phrase 
$\II(f)$ with its parameters instantiated to $\infsemFO{\II}{\formulas}{t_1},
\ldots,\infsemFO{\II}{\formulas}{t_n}$)}
\end{minipage}\rule{0pt}{0.55cm}\vspace{0.0cm}\\
\hline
\hspace{-0.3in}$P(t_1,\ldots,t_n)$ & & \begin{minipage}{8cm}$\II(P)\langle \infsemFO{\II}{\formulas}{t_1},\ldots,\infsemFO{\II}{\formulas}{t_n}\rangle$\\\emph{(i.e., the declarative sentence
$\II(P)$ with its parameters instantiated to $\infsemFO{\II}{\formulas}{t_1},
\ldots,\infsemFO{\II}{\formulas}{t_n}$)}
\end{minipage}\rule{0pt}{0.55cm}\vspace{0.0cm}\\
\hline
\hspace{-0.3in}$\varphi \lor \psi$ & & $\infsemFO{\II}{\formulas}{\varphi}$ or
$\infsemFO{\II}{\formulas}{\psi}$ (or both)\rule{0pt}{0.2cm}\vspace{0.0cm}\\
\hline
\hspace{-0.3in}$\varphi \land \psi$ & & $\infsemFO{\II}{\formulas}{\varphi}$ and
$\infsemFO{\II}{\formulas}{\psi}$\rule{0pt}{0.2cm}\vspace{0.0cm}\\
\hline
\hspace{-0.3in}$\lnot \varphi$ & & \begin{minipage}{8cm}
it is not the case that $\infsemFO{\II}{\formulas}{\varphi}$\\
\emph{(i.e., $\infsemFO{\II}{\formulas}{\varphi}$ is false)}\end{minipage}\rule{0pt}{0.55cm}\vspace{0.0cm} \\
\hline
\hspace{-0.3in}$\varphi \mim\psi$ & & \begin{minipage}{8cm}
if $\infsemFO{\II}{\formulas}{\varphi}$ then
$\infsemFO{\II}{\formulas}{\psi}$\newline \emph{(in the sense of material
implication)}\end{minipage}\rule{0pt}{0.55cm}\vspace{0.0cm}\\
\hline
\hspace{-0.3in}$\exists x\ \varphi$ & & \begin{minipage}{8cm}
there exists an $x$ in the universe of discourse such that
$\infsemFO{\II}{\formulas}{\varphi}$\end{minipage}\rule{0pt}{0.55cm}\vspace{0.0cm}\\
\hline
\hspace{-0.3in}$\forall x\ \varphi$ & & for all $x$ in the universe of discourse, $\infsemFO{\II}{\formulas}{\varphi}$\rule{0pt}{0.2 cm}\vspace{0.0cm}\\
\hline
\hspace{-0.3in}$t_1=t_2$ & & \begin{minipage}{8cm}
$\infsemFO{\II}{\formulas}{t_1}$ and $\infsemFO{\II}{\formulas}{t_2}$ are the same\\\emph{(i.e., they represent the same elements of the universe 
of discourse)}\end{minipage}\rule{0pt}{0.55cm}\vspace{0.0cm}\\
\hline
\hspace{-0.3in}$T = \{\varphi_1,\ldots,\varphi_n\}$ & & $\infsemFO{\II}{\formulas}{\varphi_1}$ and \ldots
and $\infsemFO{\II}{\formulas}{\varphi_n}$\rule{0pt}{0.2cm}\vspace{0.0cm} \\
\hline\hline
\end{tabular}
\end{center}
\end{table}

A special case in Table \ref{fig:FOinf:form} is its final row.
It gives the meaning of the \emph{implicit} composition operator of FO, i.e., 
the operator that forms a single theory out of a number of sentences. 
The informal semantics of this operator is simply \cred{that of the} 
standard (monotone) conjunction. 

\begin{example} For the theory $T_{col}$ defined in Example~\ref{ex:col},
$\infsemFO{\II_{col}}{\formulas}{T_{col}}$  results in the following statement:
\begin{quote}
\textcolor{black}{
For all $x$ in the universe of discourse, for all $y$ in the universe of 
discourse, if there is an edge from $x$ to $y$, then it is not the case 
that the color of $x$ and the color of $y$ are the same.}
\end{quote}
In other words, $T_{col}$ states that adjacent nodes are of different color.
\end{example}

\textcolor{black}{
\begin{example} \label{ex:sibling}
Let us consider an alternative application 
domain for the FO language considered above, in which we have humans, each of 
some age, and each possiby with some siblings. We now may have the following 
intended interpretation $\II_{sib}$ of the vocabulary $\Voc_{col}$:
	\begin{itemize} 
	\item $\II_{sib}(Node/1) =$``$\#_1$ is a human''; 
	\item $\II_{sib}(Edge/2) =$``$\#_1$ and $\#_2$ are siblings'';
        \item $\II_{sib}(\mathit{ColorOf}/1) =$``the age of $\#_1$''.
	\end{itemize}
For theory $T_{col}$,
$\infsemFO{\II_{sib}}{\formulas}{T_{col}}$ yields this statement:
\begin{quote}
For all $x$ in the universe of discourse, for all $y$ in the universe 
of discourse, if $x$ and $y$ are siblings, then it is not the case that 
the age of $x$ and the age of $y$ are the same. 
\end{quote}
In other words, in the ``sibling'' application domain, the theory $T_{col}$
states that siblings have different age.
\end{example}}


\paragraph{Informal semantics for FO's semantical concepts}

In addition to explaining the informal meaning of syntactical
expressions, the informal semantics of FO also offers explanations 
of FO's formal semantical objects:  structures, the satisfaction relation, 
and the derived concepts of entailment, satisfiability, and  validity. 


\textcolor{black}{The basic informal notion behind these concepts is that of a {\em state 
of affairs}. 
States of affairs differ in the objects that exists in 
the {\em universe of discourse}, or in the relationships and functions amongst 
these objects. The application domain is in one of many potential
{states of affairs}. In a state  of affairs, a proposition of the application domain is either true or false.} 


\textcolor{black}{
The intended interpretation $\II$ in general does not fix the state of affairs. 
Rather, it determines an abstraction function from states of affairs 
to $\Voc$-structures.\footnote{More accurately, it 
determines an abstraction function from states of affairs to classes of 
isomorphic $\Voc$-structures.}}

\newcommand{\stateOA}{{\mathcal S}}


\textcolor{black}{
\begin{example}
Under the intended interpretation $\II_{col}$ for the vocabulary $\Voc_{col}$,
the $\Voc_{col}$-structure $\I_{col}$ of Example~\ref{ex:col:structure} 
represents any state of affairs with five elements in the universe of
discourse: three nodes abstracted as $n_1, n_2, n_3$ with edges corresponding 
to $(n_1,n_2)$ and $(n_2,n_3)$ and two colors represented by $c_1, c_2$; 
finally, a coloring mapping that associates colors to all elements (nodes 
and colors). 
\end{example}}

In the sequel, we denote the class of states of affairs that abstract under $\II$ to structure $\I$ as $\infsemFO{\II}{\structures}{\I}$. We call this the informal semantics of the structure $\I$ under $\II$.  Table~\ref{fig:FOinf:struct} expresses the meaning of structures as explained above.

\textcolor{black}{
Different intended interpretations $\II$ give rise to different abstractions.
\begin{example}
Under the alternative intended interpretation $\II_{sib}$ of
Example~\ref{ex:sibling}, $\I_{col}$ represents any state of affairs with
three persons and two ages, where the sibling relation consists of pairs 
corresponding to  $(n_1,n_2)$ and $(n_2,n_3)$ and where persons $n_1, n_3$ 
have the same age different from that of $n_2$. However, no {\em possible} 
state of affairs under this intended interpretation abstracts into $\I_{col}$,
since the sibling relation amongst a group of persons is always an equivalence 
relation while $Edge^{\I_{col}}$ is not. Stated differently, 
$\infsemFO{\II_{sib}}{\structures}{\I_{col}}$ contains only impossible 
states of affairs.
\end{example}}

\begin{table}
\caption{The informal semantics of FO structures.\label{fig:FOinf:struct}}
\textcolor{black}{
\begin{center}
\begin{tabular}{ccl}
\hline\hline
$\I$ & \phantom{aaaaa} & a state of affairs $\stateOA\in\infsemFO{\II}{\structures}(\I)$ that has abstraction $\I$ \rule{0pt}{0.2cm}\\
\hline
$\dom{\I}$ & & the set of elements in the universe of discourse of $\stateOA$\\ 
\hline
$P^\I $ \rule{0pt}{0.2cm}& & the property of $\stateOA$ described by the declarative sentence $\II(P)$\\
\hline
$f^\I$\rule{0pt}{0.2cm} & & the function in $\stateOA$ described by the noun phrase $\II(f)$\\
\hline\hline
\end{tabular}
\end{center}
}
\end{table}

\textcolor{black}{
The informal semantics of the  satisfaction relation $\models$ between structures and sentences is the relation between states of affairs and true propositions in it. That is, $\I\models\varphi$ is interpreted as
``{\em (the proposition represented by) $\varphi$ is true in the state of
affairs (corresponding to) $\I$.}'' Table
\ref{fig:FOinf:struct2} summarizes this observation.}

\begin{table}
\caption{The informal semantics of the satisfaction relation
of FO.\label{fig:FOinf:struct2}}
\begin{center}
\begin{tabular}{ccl}
\hline\hline
$\models$ & \phantom{aaaaa} & \hspace{1.4in} $\infsemFO{\II}{\models}$ \\
\hline\hline
$\I \models T$ & & The property $\infsemFO{\II}{\formulas}{T}$ holds in
the states of affairs $\infsemFO{\II}{\structures}{\I}$ \rule{0pt}{0.4cm}\\
\hline\hline
\end{tabular}
\end{center}
\end{table}

We thus specified for each vocabulary $\Voc$ and for every intended 
interpretation $\II$ for $\Voc$ a triple $\infsemFO{\II} = 
(\infsemFO{\II}{\formulas{\Voc}},\infsemFO{\II}{\structures{\Voc}},
\infsemFO{\II}{\models})$ that explains the informal semantics of formulas, 
structures and the satisfaction relation. We call this the {\em (standard) 
theory of informal semantics of FO}.



\paragraph{Informal semantics of derived semantical relations}

The standard theory of informal semantics of FO also 
induces the informal meaning for the derived semantical concepts of 
entailment, equivalence, validity, satisfiability, and equivalence 
in a way that reflects our understanding of these concepts 
in mathematics and formal science.

For instance, under the standard informal semantics of FO, a formal 
expression $\psi\models\varphi$ (entailment), becomes the statement that 
the informal semantics of $\varphi$ (a statement that expresses a property 
of the application domain)  is true in every state of affairs for that application domain, in which the informal semantics of $\psi$ (another statement that expresses a 
property of the application domain)  is true (which, in case the informal semantics of $\psi$ and $\varphi$ are
mathematical propositions, means that the first proposition, the one
corresponding to $\psi$, mathematically entails the second proposition,
the one corresponding to $\varphi$). Similarly, validity of $\varphi$ means that the informal semantics 
of $\varphi$ is true in every state of affairs, and satisfiability of 
$\varphi$ means that the informal semantics of $\varphi$ is true in at 
least one state of affairs.

\paragraph{Precision of the informal semantics}



\textcolor{black}{
The informal semantics $\infsemFO{\II}{\formulas}(\varphi)$ of a sentence 
of theory $T$ under intended interpretation $\II$ is a syntactically 
correct statement in natural language, but is this statement a sensible and 
precise statement about the application domain? This is a concern 
given the vagueness, ambiguity and the lack of coherence that is so often 
ascribed to  natural language \citep{BarwiseCooper81}. We now discuss this.}

\textcolor{black}{
First, when the intended interpretation $\II$ of $\Voc$ is vague or ambiguous,
then indeed, the informal semantics of FO sentences over $\Voc$ will be vague
or ambiguous. E.g., if we interpret $Edge^\II$ as "$\#_1$ is rather similarly 
looking as $\#_2$", then surely the informal semantics of sentences containing 
this predicate will be vague. FO is not designed to express information given 
in terms of vague concepts. In FO, as in many other languages, it is the 
user's responsibility to design a vocabulary with a clear and precise intended 
interpretation.}

A second potential source of ambiguity lies in the use of natural language connectives such as ``and'', ``or'', ``if\dots then\dots'', etc. in Table~\ref{fig:FOinf:form}. Several of these connectives are overloaded, in the sense that they may mean different things in different contexts. 
However, this does not necessarily lead to ambiguity or vagueness, since human readers are skilled in using context to disambiguate overloaded concepts. Let us consider several potential ambiguities.


In natural language, the connective ``and'' corresponds not only to logical 
conjunction, but also to temporal consecutive conjunction exemplified in 
{\em I woke up and brushed my teeth}. However, the temporal interpretation 
arises only in a temporal context. Table~\ref{fig:FOinf:form} is 
\textcolor{black}{not stated within a temporal context, so it is logical 
conjunction that is intended and this intended meaning is inherited in 
every occurrence of $\land$ in every FO sentence.} Similarly, the word 
``or'' is used to denote  both inclusive and exclusive disjunction. 
\textcolor{black}{In mathematical texts its accepted meaning is that of 
inclusive disjunction. The rule for ``or'' in Table~\ref{fig:FOinf:form} 
explicitly adds ``or both'', to remove any possibility for ambiguity.}


The conditional ``if \dots then \dots''  is famously ambiguous. It can mean 
many different things in varying contexts \citep{DancyS2005}. Therefore, 
{\em any} choice for the formal and informal semantics of the implication 
symbol can only cover part of the use of the conditional in natural 
language\footnote{\textcolor{black}{This suggests adding symbols to the logic
to express other conditionals, which is something we we will do later in this
paper).}}. 
In FO, the informal semantics of $\Rightarrow$ was chosen to be the material 
implication, which interprets ``if A then B''  as ``not A or B''. It has the 
benefit of being simple and clear, \textcolor{black}{and it is likely the 
conditional that we need and use most frequently in mathematical text.} 

To summarize, the natural language connectives and quantifiers in  Table~\ref{fig:FOinf:form} are precise  and clear. This precision  is inherited by the informal semantics $\infsemFO{\II}{\formulas}$ of FO sentences. Consequently, under the assumption that the intended interpretation $\II$ is also clear, the natural language statements produced by $\infsemFO{\II}{\formulas}$ are as clear and unambiguous as mathematical text.

\paragraph{Informal semantics as the ``empirical'' component of logic}

Formal sentences of a knowledge representation logic are used to specify information about an application domain. 
The role of the informal semantics theory of a knowledge representation logic is  to provide a principled account of which information about an application domain is expressed by formal sentences of the logic. In other formal empirical sciences, we find theories with a similar role. A physics theory (e.g., quantum mechanics) not only consists of mathematical equations but, equally important, also of a theory that describes, often with extreme precision, how the mathematical symbols used in these equations are linked to observable phenomena and measurable quantities in reality. This second part of the theory therefore plays a role similar to that of the informal semantics theory of a knowledge representation logic. 

\textcolor{black}{
A physics theory is a precise, falsifiable hypothesis about the reality it 
models. Such a theory can never be proven. But it potentially can be 
experimentally falsified by computing a mathematical prediction from the 
formal theory, and verifying if the measured phenomena match the predictions. 
If not, the experiment refutes the theory. Otherwise, it corroborates (supports)
the theory.} Availability of a large and diverse body of corroborating 
experiments, in the absence of experimental refutation, increases our 
confidence in the theory.


Likewise, a formal logic with a theory of informal semantics is a precise, 
falsifiable hypothesis about the forms of  information (the ``thoughts'') 
expressible in the logic and their truth in different states of affairs. 
\textcolor{black}{A possible experiment here 
could consist of a choice of a vocabulary $\Voc$ 
with intended interpretation $\II$, a $\Voc$-theory $T$ and a $\Voc$-structure 
$\I$}. The experiment would falsify the theory of informal semantics if 
$\I \models T$ and yet $\infsemFO{\II}{\formulas}{T}$ would not be considered
to be true in the states of affairs of $\infsemFO{\II}{\structures}{\I}$ by 
human reasoners. Or, conversely, if it holds that $\I \not\models T$ while
$\infsemFO{\II}{\formulas}{T}$ would be considered by human reasoners to
be true in $\infsemFO{\II}{\structures}{\I}$. 
{That such mismatches do not occur corroborates the theory of informal
semantics of FO. Here, we present one such corroborating experiment.}

\begin{example}
Let us return to our running
example. Since $\I_{col} \models T_{col}$, we  check that
\cred{$\infsemFO{\II}{\models}(\I_{col},T_{col})$}\ignore{$\infsemFO{\II}{\models}(\I_{col} \models T_{col})$} also holds. \cred{Indeed, it
is evident that human reasoners would determine that in all states-of-affairs 
described by $\infsemFO{\II}{\structures}(\I_{col})$ it is the case that $\infsemFO{\II}{\formulas}(T_{col})$} (all neighbouring nodes have different colors).
Therefore, this experiment
corroborates the standard theory of FO's informal semantics.
\end{example}

\paragraph{\textcolor{black}{Alternative theories of informal semantics of FO}}
\textcolor{black}{That the informal and formal semantics correspond so well 
to each other is not an accident. The key to this is the tight correspondence 
between the natural language expressions occurring in 
Table~\ref{fig:FOinf:form} and the statements used in the bodies of 
rules of Definition~\ref{deftrue}, for instance, between the informal and 
formal semantics of $\lor$, where we have 
{``{\bf $\infsemFO{\II}{\formulas}{\varphi}$ {{or} $\infsemFO{\II}{\formulas}{\psi}$ {(or both)}}},'' on the one side, and the condition for 
$\I\models\psi\lor\varphi$, ``{\bf $\I\models  \psi$ {or} $\I\models 
\varphi$ {(or both)}},'' on the other.}  Breaking this correspondence would lead to many mismatches between informal and formal semantics.  E.g., suppose we modify  the formal semantics:
\begin{center} $\I\models \psi\lor\varphi$ if $\I\models\psi$ or (exclusively) $\I\models\varphi$
\end{center} but we keep the entry for $\lor$ in Table~\ref{fig:FOinf:form}. That would obviously create  a range of experiments in which formal and informal semantics are in mismatch. E.g., for the propositional vocabulary $\Voc=\{P,Q\}$, it now holds that $\{P,Q\}\not\models P\lor Q$ while under any intended interpretation of $P, Q$, the statement "$\infsem{\II}(P)$ or $\infsem{\II}(Q)$ (or both)" is true in the state of affairs $\infsemFO{\II}{\structures}(\{P,Q\})$.
}

\textcolor{black}{
{However, while the formal semantics constrains the informal semantics, it does not {\em uniquely} determine it.} The informal semantics of formulas arises from a coherent set of 
interpretations of language constructs and of the semantic primitives: 
structures and $\models$. By carefully changing this system, we may obtain 
an informal semantics that still matches with the formal semantics although 
it assigns a very different meaning to formulas. Let us illustrate this. }


\textcolor{black}{
In the proposed new informal semantics, we reinterpret all connectives such 
that the informal semantics of any sentence is exactly the negation of its 
standard informal semantics, and we reinterpret  $\I\models \varphi$ to mean 
that the informal semantics of $\varphi$ is {\em false} in states of affairs 
represented by $\I$. These two negations then compensate for each other, leading to a theory of informal semantics that still matches with the formal 
semantics, even though it assigns the negation of the standard informal 
semantics to each FO sentence! To be precise, the alternative informal 
semantics theory is given by}
$(\infsemFOA{\II}{\formulas},\infsemFOA{\II}{\structures},
\infsemFOA{}{\models})$, where $\infsemFOA{\II}{\structures}=\infsemFO{\II}{\structures}$ that is, the informal semantics of structures as states of affairs 
is as in the standard informal semantics of FO, and
the interpretations of the two non-standard components 
$\infsemFOA{\II}{\formulas}$ and $\infsemFOA{}{\models}$ are defined in 
Table \ref{fig:FOAinf}.  To illustrate, let us consider the formula 
$\hat{\varphi}_{col}$:
\[
\exists x\exists y\ (Edge(x,y) \land \mathit{ColorOf}(x) = \mathit{ColorOf}(y)).
\]
The non-standard informal semantics
$\infsemFOA{\II}{\formulas}$ interprets the formula
$\hat{\varphi}_{col}$ as:
\begin{quote}
\textcolor{black}{For all $x$ in the universe of discourse, for all $y$ in 
the universe of discourse it is not the case that there is an edge from $x$
to $y$, or the color of $x$ and the color of $y$ are not the same.}
\end{quote}
\textcolor{black}{Stated differently, $\infsemFOA{\II}{\formulas}$ says that adjacent nodes are of different color.
Note that this statement is the negation of the standard informal semantics 
of this formula and that it has the same meaning as the one produced by
the standard informal semantics $\infsemFO{\II_{col}}{\formulas}$ for the 
(different) formula $T_{col}$. 
Since in the new informal semantics, structures are still interpreted in 
the same way as before, it follows that 
$\infsemFOA{\II_{col}}{\formulas}(\hat{\varphi}_{col})$ is satisfied in 
the state-of-affairs $\infsemFOA{\II_{col}}{\structures}{{\I}_{col}}$. 
On the formal side, nothing has changed and it is the case that 
$\I_{col}\not\models \hat{\varphi}_{col}$, i.e., the formula is not 
formally satisfied in the structure. But under the new informal semantics, 
the relation $\models$ is now interpreted as non-satisfaction and hence, 
$\I_{col}\not\models \hat{\varphi}_{col}$ is to be interpreted as the fact 
that $\infsemFOA{\II_{col}}{\formulas}(\hat{\varphi}_{col})$ is 
\mbox{(not non-)satisfied} in $\infsemFOA{\II_{col}}{\structures}{{\I}_{col}}$. 
Which is true!}

%

\textcolor{black}{
Thus, even though the formal semantics of a 
logic strongly constrains its informal semantics, there may nevertheless 
remain different informal semantics that correspond to the logic's formal 
semantics. 
{In the case of FO, an informal semantics such as $\infsemFOA{\II}{\formulas}$ 
is counterintuitive and 
of no 
practical use. 
}}  

\begin{table}[ht]
\caption{A non-standard informal semantics of FO. The informal semantics of 
terms is as in Table \ref{fig:FOinf:form} and the informal 
semantics of structures as in Table \ref{fig:FOinf:struct}.
\label{fig:FOAinf}}
\begin{center}
\begin{tabular}{ccp{8cm}}
\hline\hline
\hspace{-0.3in}$\Phi$ & \phantom{aaaaa} & \hspace{1.3in}$\infsemFOA{\II}{\formulas}{\Phi}$\rule{0pt}{0.2cm}\vspace{0.0cm}\\
\hline\hline
\hspace{-0.3in}$P(\vec{t})$ & & 
it is not the case that $\II(P)\langle \infsemFOA{\II}{\formulas}{t_1},\ldots,\infsemFOA{\II}{\formulas}{t_n}\rangle$ 
\rule{0pt}{0.2cm}\vspace{0.0cm}\\
\hline
\hspace{-0.3in}$\varphi \lor \psi$ & & $\infsemFOA{\II}{\formulas}{\varphi}$ and
$\infsemFOA{\II}{\formulas}{\psi}$\rule{0pt}{0.2cm}\vspace{0.0cm}\\
\hline
\hspace{-0.3in}$\varphi \land \psi$ & & $\infsemFOA{\II}{\formulas}{\varphi}$ or
$\infsemFOA{\II}{\formulas}{\psi}$\rule{0pt}{0.2cm}\vspace{0.0cm}\\
\hline
\hspace{-0.3in}$\lnot \varphi$ & & it is not the case that $\infsemFOA{\II}{\formulas}{\varphi}$\rule{0pt}{0.2cm}\vspace{0.0cm}\\
\hline
\hspace{-0.3in}$\exists x\ \varphi$ & & for all $x$ in the universe of discourse,
$\infsemFOA{\II}{\formulas}{\varphi}$\rule{0pt}{0.2cm}\vspace{0.0cm}\\
\hline
\hspace{-0.3in}$\forall x\ \varphi$ & & there exists an $x$ in the universe 
of discourse such that $\infsemFOA{\II}{\formulas}{\varphi}$\rule{0pt}{0.2cm}\vspace{0.0cm}\\
\hline
\hspace{-0.3in}$t_1=t_2$ & & 
$\infsemFOA{\II}{\formulas}{t_1}$ and $\infsemFOA{\II}{\formulas}{t_2}$ are not the same \rule{0pt}{0.2cm}\vspace{0.0cm}\\
\hline
\hspace{-0.3in}$T = \{\varphi_1,\ldots,\varphi_n\}$ & & $\infsemFOA{\II}{\formulas}{\varphi_1}$
or \ldots
or $\infsemFOA{\II}{\formulas}{\varphi_n}$\rule{0pt}{0.2cm}\vspace{0.0cm}\\
\hline\hline
& &  \\
\hline\hline
\hspace{-0.3in}$\models$ & & \hspace{1.4in}$\infsemFOA{}{\models}$ \\
\hline\hline
\hspace{-0.3in}$\I \models T$ & & \begin{minipage}{8cm}
The property $\infsemFOA{\II}{\formulas}{T}$ does not hold in 
the state-of-affairs  $\infsemFO{\II}{\structures}{\I}$
\end{minipage}\rule{0pt}{0.3cm}\vspace{0.0cm}\\
\hline\hline
\end{tabular}
\end{center}
\end{table}


As many 
logics reuse the connectives of FO, we will use the following terminology. 
We will say that a connective 
is \emph{classical} in a logic under some informal semantics $\infsem{\II}$ 
if $\infsem{\II}{\formulas}$ interprets it in the same way
as $\infsemFO{\II}{\formulas}$ does (i.e., by the
same natural language phrase). For
instance, in the non-standard informal semantics
$\infsemFOA{\II}{\formulas}$ for FO, the negation connective  $\lnot$
 is classical, whereas the conjunction connective $\land$
is not. 

In the rest of this paper, we often omit the superscript from the notation $\infsem{\II}{\formulas},\infsem{\II}{\structures},\infsem{\II}{\models}$ when it is clear which of the three components of a theory of informal semantics $\infsem{\II}$ is intended.

\section{The 
original formal and informal semantics of extended logic programs} \label{sec:infsemGL}


In this section, we recall the standard formal and informal semantics of 
extended logic programs as it appeared in \cite{ge1,ge2}. An alternative review of the informal semantics for this logic can be found in~\cite[Section 2.2.1]{gelkahl14}. 
\cred{For simplicity, here we only consider the propositional case. 	
	Logic programs with variables are interpreted by the means of their so-called grounding that transforms them into propositional programs.} 
	
A \emph{literal} is either an atom $A$ or an expression $\neg A$, where $A$ is an atom. An extended logic programming rule is an expression of the form
\beq A \rul B_1, \dots, B_n, \naf C_1, \dots,
\naf C_m,
\eeq{eq:rule}
where $A$, $B_i$, \cred{and} $C_j$ are propositional literals.  If all the literals \cred{$A$,} $B_i$ and $C_i$ are atoms (i.e., symbol $\neg$ does not appear in the rule), then such a rule is called {\em normal}. 
 The \cred{literal}\ignore{atom} $A$ is the \emph{head} of the rule and expression $B_1,\ldots,B_n, \naf C_1, \dots, \naf C_m$ is its \emph{body}. We may abbreviate this rule as $Head\rul Body$.
An \emph{extended logic program} is a
finite set of such rules. We denote the set of all programs in a given
vocabulary $\Voc$ by $\mathbb{P}_{\Voc}$.

A {\em consistent} set of propositional literals is  a set that does not contain
both  $A$ and its {\em complement} ${\neg}A$ for any atom $A$.
A \emph{believed literal set} $X$ is a consistent set of propositional literals. 
We denote the set  of
all believed literal sets for a vocabulary $\Voc$ by $\mathbb{B}_{\Voc}$. A believed literal set $X$
\emph{satisfies} a rule $r$ of the form~\eqref{eq:rule} if $A$ belongs to $X$ or
there exists an $i \in \{1,\ldots,n\}$ such that $B_i \not\in X$ or a $j \in
\{1,\ldots,m\}$ such that $C_j \in X$. A believed literal set is a {\em model} of a program $P$ if
it satisfies all rules $r \in P$.

For a rule
$r$ of the form~\eqref{eq:rule} and a believed literal set $X$,  
the reduct $r^X$  is defined whenever there is
no literal $C_j$ for $j \in
\{1,\ldots,m\}$ such that $C_j \in X$. If the reduct $r^X$ is defined, then it
is the rule 
\[
A \rul B_1,\ldots,B_n.
\]
The reduct $P^X$ of the program $P$ consists of the rules $r^X$ for all 
$r\in P$, for which the reduct is defined.
A believed literal set $X$ is a \emph{stable model} or \emph{answer set} of $P$,
denoted $X \models_{st} P$, 
if it is a $\subseteq$-least model of $P^X$. 

The formal logic of extended logic programs consists of the triple 
$(\mathbb{P}_{\Voc},\mathbb{B}_{\Voc},\models_{st})$.
Gelfond and Lifschitz~(\citeyear{ge1,ge2}) described an informal semantics for such programs based on \emph{epistemic} notions of default and autoepistemic reasoning. Just as with $\infsemFO{\II}$, the informal semantics $\infsemGL{\II}$ arises from a system of interpretations of the formal syntactical and semantical concepts. 
We now recall this informal semantics, which we name $\infsemGL{\II}$. 
We 
denote its three components by $\infsemGL{\II}{\formulas}$, $\infsemGL{\II}{\structures}$ and $\infsemGL{\II}{\models}$.

\textcolor{black}{ 
One of the key aspects of $\infsemGL{\II}$ is that it views a believed literal set $X$ as an abstraction of a \emph{belief state} of some agent.  An agent in some belief state considers certain states of affairs as possible and the others as impossible. \textcolor{black}{The corresponding believed literal set $X$ is the set of all literals $L$ that the agent believes in, that is, those that are true in all states of affairs that agent regards as possible. Importantly, it is not the case that
a literal $L$ that does not belong to $X$ is believed to be false by the agent.}
Rather, it is \emph{not believed} by the agent: the literal is false in some states of affairs the agent holds possible, and it may be true in others. In fact, $L$ is true in at least one of the agents possible state of affairs unless the complement of $L$ belongs to $X$ (unless the agent believes the complement of $L$). Thus,  the informal semantics $\infsemGL{\II}$ explains the meaning of programs in terms of what literals an agent with incomplete knowledge of the application domain might believe in. 
}

\ignore{
\textcolor{blue}{
One of the key aspects of $\infsemGL{\II}$ is that it views an answer set
(formally, a certain belief set) as an abstraction of a \emph{belief state}, 
that is, a collection of states of affairs the agent holds possible. Thus, 
this setting is designed to explain the meaning of programs in terms of what
an agent with incomplete knowledge of the application domain believes in.
As in the informal semantics for FO, under $\II$, each literal in a belief 
set corresponds to a property of the application domain. However, the 
presence of such a literal in a belief set does not express that this 
property \emph{objectively} holds in the application domain. Rather, it 
says that the agent \emph{believes} the property, that is, the property
holds in all state of affairs the agent holds possible. Further, a literal
that does not show in a belief set is not false. Rather, it is not believed 
by the agent. This is in contrast with objective informal semantics, such 
as the one we discussed for FO, where truth of a formula in a model simply
means that the corresponding informal proposition holds in the state of affairs
of the aplicaton domain.}    
}

\begin{example}
Consider the believed literal set \[X=\{student(mary), male(john)\}\] under the obvious intended interpretation $\II$ for the propositional atoms. This $X$ is the abstraction of any belief state in which the agent both believes that Mary is a student and that John is male, and does not believe that John is a student or that Mary is male. One such belief state is the state $B_0$ in which the agent considers the following states of affairs as possible:
\begin{enumerate}
\item John is the only male in the domain of discourse; Mary is the only student.
\item\label{it:2} John and Mary are both male students.
\item\label{it:3} John and Mary are both male; Mary is the only student.
\item\label{it:4} John is the only male; John and Mary are both students
\end{enumerate}
\textcolor{black}{Another belief state corresponding to $X$ is the state $B_1$ 
in which the agent considers the states of affairs~\ref{it:2}-\ref{it:4} of $B_0$ as possible. 
In both, Mary is a student and John a male in all possible states of affairs. John is student in worlds \ref{it:2},\ref{it:4}; Mary is male in worlds \ref{it:2},\ref{it:3}. Hence, literals $\neg student(john)$ and $\neg male(mary)$ are not believed.}

\textcolor{black}{Although both belief states abstract to the same believed literal set, they are different. 
When compared to $B_0$, the belief state $B_1$ contains the additional belief that either John is a student or Mary is male. Since this additional belief is not atomic, a formal believed literal set cannot distinguish belief states in which it holds (e.g., $B_1$) from belief states in which it does not (e.g., $B_0$).  In this way, different informal belief states are still abstracted to the same formal believed literal set. }\textcolor{black}{This shows that believed literal sets are a rather \emph{coarse} way of abstracting belief states, compared to, e.g., believed formula sets or Kripke structures. }   
\end{example}

\begin{table}
\caption{The Gelfond-Lifschitz informal semantics of belief sets.\label{fig:GLinf:struc}}
\begin{center}
\begin{tabular}{ccl}
\hline\hline
A belief set $X$ & \phantom{aaaaa} & A belief state $B \in\infsemGL{\II}{\structures}(X)$ that has abstraction $X$ \rule{0pt}{0.4cm}\\
\hline
$A \in X$ for atom $A$ & & \begin{minipage}{8cm}$B$ has the belief that $\II(A)$ is true;\\ i.e., $\II(A)$ is true in all states of affairs possible in $B$ \end{minipage}\\
\hline
$\lnot A \in X$ for atom $A$ & &  \begin{minipage}{8cm}
$B$ has the belief that $\II(A)$ is false;\\ i.e., $\II(A)$ is false in all states of affairs possible in $B$\end{minipage}\\
\hline
$A \not\in X$ for atom $A$ & &\begin{minipage}{8cm} $B$ does not have the belief that $\II(A)$ is 
true;\\ i.e., $\II(A)$ is false in some state of affairs possible in $B$\end{minipage}\\
\hline
$\lnot A \not\in X$ for atom $A$ & & \begin{minipage}{8cm}$B$ does not have the belief that $\II(A)$ is false;\\ i.e., $\II(A)$ is true in some state of affairs possible in $B$\end{minipage}\\
\hline\hline
\end{tabular}
\end{center}
\end{table}

We denote the class of informal belief states that are abstracted  to a given  formal believed literal set $X$ under an intended interpretation $\II$ as $\infsemGL{\II}{\structures}{X}$. Table \ref{fig:GLinf:struc} summarizes this abstraction function. 

Table \ref{fig:GLinf:form} shows the Gelfond-Lifschitz informal semantics $\infsemGL{\II}{\formulas}$ of programs. As in the informal semantics for FO, each atom $A$ has an intended interpretation $\II(A)$ which is represented  linguistically as a noun phrase about the application domain. The intended interpretation $\II(\neg A)$ is ``it is not the case that $\II(A)$''.   As is clear from this table, under $\infsemGL{\II}{\formulas}$, extended logic programs have both classical and non-classical
connectives. On the one hand, the comma operator is classical conjunction and the rule
operator $\rul$ is classical implication. On the other hand, the implicit composition
operator (constructing a program out of individual rules) is non-classical, because it performs a closure operation:
the agent knows \emph{only} what is explicitly stated.  Of the two
negation operators, symbol $\neg$ is classical negation, whereas {\tt not} is
a non-classical negation, which is called \emph{default negation}.

\begin{table}
\caption{The Gelfond-Lifschitz (\citeyear{ge1,ge2}) informal semantics for ASP formulas.\label{fig:GLinf:form}}
\begin{center}
\begin{tabular}{lcp{6cm}}
\hline
\hspace{2cm}$\Phi$  & \phantom{aaaaa} & \rule{0pt}{0.2cm}\hspace{2cm} $\infsemGL{\II}{\formulas}{\Phi}$\\
\hline\hline
propositional atom $A$ & & { $\II(A)$ } \rule{0pt}{0.2cm}\\ 
\hline			
propositional literal ${\neg}A$ & & it is not the case that $\II(A)$
\rule{0pt}{0.2cm}\\ 
\hline 
expression of the form {\tt not~}$C$ & & the agent does not know that $\infsemGL{\II}{\formulas}{C}$\rule{0pt}{0.2cm}\vspace{0.0cm}\\ 
\hline 
expression of the form $\Phi_1,\Phi_2$ & & $\infsemGL{\II}{\formulas}{\Phi_1}$ and $\infsemGL{\II}{\formulas}{\Phi_2}$ \rule{0pt}{0.2cm}\vspace{0.0cm}\\
\hline
rule $Head\rul Body$ & & \begin{minipage}{7cm}
if $\infsemGL{\II}{\formulas}{Body}$ then $\infsemGL{\II}{\formulas}{Head}$\\
\emph{(in the sense of material implication)}\end{minipage}\rule{0pt}{0.2cm}\vspace{0.0cm}\\
\hline
program $P=\{r_1,\ldots,r_n\}$ & & \begin{minipage}{6cm}
	\rule{0pt}{0.2cm}  All the agent knows is:
	\vspace{-\topsep} 
	\begin{itemize}
					\setlength{\itemsep}{0pt}
					\setlength{\topsep}{0pt} 
					\setlength{\parsep}{0pt} 
					\setlength{\parskip}{0pt}
					\setlength{\partopsep}{0pt}
					\item $\infsemGL{\II}{\formulas}{r_1}$ and \rule{0pt}{0.4cm}
					\item \ldots
					\item $\infsemGL{\II}{\formulas}{r_n}$ \rule{0pt}{0.4cm}
				\end{itemize}
			\end{minipage}\vspace{0.0cm}\\
			\hline\hline
		\end{tabular}
	\end{center}
\end{table}


The final component of the GL theory of informal semantics is 
$\infsemGL{\II}{\models_{st}}$, which explains what it means for a set 
of literals $X$ to be a stable model of a program $P$. As can be seen 
in Table \ref{fig:GLinf:models}, this means that, given that 
$\infsemGL{\II}{\formulas}{P}$ 
represents
precisely the knowledge of the agent, $X$ could be the set of literals the
agent believes.

\begin{table}
\caption{The Gelfond-Lifschitz (\citeyear{ge1,ge2}) informal semantics for the ASP satisfaction relation.\label{fig:GLinf:models}}
\begin{tabular}{lcp{9cm}}
\hline\hline
\hspace{0.5cm}$\models_{st}$ & \phantom{aaaaa}& \hspace{2cm}$\infsemGL{\II}{\models_{st}}$\\
\hline\hline
	$X\models_{st} P$ & & \rule{0pt}{0.4cm}Given that all
        the agent knows is $\infsemGL{\II}{\formulas}{P}$, $X$ could be the 
        set of literals the agent believes\\
			\hline\hline
                      \end{tabular}
                    \end{table}


To illustrate this informal semantics, let us consider Gelfond and Lifschitz' 
well-known \emph{interview} example. 
\begin{example}


\newcommand{\ivu}{\mathit{Interview}}

Whether students of a certain school are eligible for a scholarship
depends on their GPA and on their minority status. The
school has an incomplete database about candidate students. Students for
which the school has insufficient information to decide eligibility
should be invited for an interview. The following ELP program expresses
the school's knowledge.
\begin{align*}
Eligible(x) &\rul HighGPA(x).\\
Eligible(x) &\rul FairGPA(x), Minority(x).\\
\lnot Eligible(x) &\rul \lnot FairGPA(x), \lnot HighGPA(x).\\
\ivu(x) &\rul {\tt not~} Eligible(x), {\tt not~} \lnot Eligible(x)\\
Minority(brit). &\\
HighGPA(mary). &\\
\lnot Minority(david). &\\
FairGPA(david). &
\end{align*}
The three rules for $Eligible$ specify a {\em partial} policy for
eligibility: they determine the eligibility for all students except for
non-minority students with fair GPA. In this sense, under
$\infsemGL{\II}$, this program does not actually \emph{define} when a
student is eligible. The next rule is \emph{epistemic}. It expresses that the 
school interviews a person whose eligibility is unknown. The remaining 
rules specify partial data on students Mary, Brit and David. In particular,
$FairGPA(brit)$ and even $FairGPA(mary)$ are unknown. 

For Mary, the first rule applies and the school
knows that she is eligible. The epistemic fourth rule will therefore not
conclude that she should be interviewed. Incidentally, nor is it implied 
that she will not be interviewed; the school (formally, the program) simply
does not know. This follows from the informal semantics of the implicit 
composition operator: ``\emph{all} the agent knows is\ldots''. For Brit 
and David, 
their eligibility is unknown. However, the reasons are different: for Brit 
because of lack of data, for David because the policy does not specify it. 
Therefore, both will be interviewed. The unique answer set extends the 
student data with the following literals:
\begin{align*}
 Eligible(mary),  \ivu(brit), \ivu(david).
\end{align*}

The crucial property of this example is that whether a student should be
interviewed does not only depend on properties of the student alone, but
also on the agent's knowledge about this student. In other words, it is
perfectly possible that the same student should be interviewed when applying
to school $A$ but not when applying to school $B$, even when the eligibility
criteria used by the two schools are precisely the same. Indeed, this can
happen if school $B$ has more information on record about the student
in question. Because of this property, a logic with a subjective informal
semantics is required here.



It is illustrative to compare this example to, for instance, graph
coloring. Whether a graph is colorable given a certain number of
colors is purely a property of this graph itself. It does not depend on
anyone's knowledge about this graph, and it does not depend on which
agent is doing the coloring. In other words, graph coloring lacks the
subjective, epistemic component of the interview example. Consequently,
as we  illustrated in Section~\ref{sec:intro}, applying $\infsemGL{\II}$
to the rules of a typical GDT graph coloring program produces a
misleading result. It refers to the knowledge of an agent, when in fact
there exists  no agent whose knowledge is relevant to the problem.

\end{example}

{The interview example, with its inherent epistemic component, is a clear case where the informal semantics $\infsemGL{\II}$ applies.} 
Similar informal semantics have also been developed for other 
formalisms for modeling reasoning with incomplete information, including 
default logic \citep{Reiter} and autoepistemic logic \citep{Moore}, and much 
of our discussion above extends to such formalisms, too. They all have a role
to play but it is important to be aware of the scope of their applicability.

\textcolor{black}{To the best of our knowledge, $\infsemGL{\II}$ is the only 
informal semantics developed in the literature for the language of extended 
logic programs. Answer set programming adopted it, when it adopted the
language of extended logic programming. However,  the
informal semantics $\infsemGL{\II}$ is not suitable for typical answer set 
programing applications. Moreover, over time, answer set programming has
developed a richer language with features not found in the original papers 
by Gelfond and Lifschitz, including choice rules, aggregates and weight 
constraints. If is often non-trivial to extend $\infsemGL{\II}$ to these richer languages, as illustrated in the next paragraph. 
}

\paragraph{Extending $\infsemGL{\II}$ to modern ASP is difficult}

{As a simple example of a modern ASP language feature, we consider 
the constraint (\ref{eq:colrule2}) used to model the property that nodes 
should be colored. 
As stated in the introduction, $\infsemGL{\II}$ provides the following informal semantics for this rule:
\textcolor{black}{``\emph{for every $x$, Aux holds if the agent does not know 
Aux and $x$ is a node and the agent does not know that $x$ has a color.}''} 

Can a simpler informal semantics (that avoids the predicate $Aux$) be given
for constraint \eqref{eq:colrule2} directly? This question is not easily 
answered. Starting from the way in which Table \ref{fig:GLinf:form} interprets 
atoms and the operator {\tt not}, one plausible candidate for such an informal 
semantics is the sentence:
``\emph{for every $x$, it is not the case that $x$ is a node and the agent does
not know that $x$ has a color}'.  

However, upon closer analysis, this sentence turns out not to match with 
the formal semantics. This can be seen as follows. Let us suppose for 
simplicity that the Herbrand universe consists of only one constant $a$. 
The proposed informal semantics boils down then to:
\begin{quote}
(*) ``\emph{It is not the case that a is a node and the agent does not know 
that a has a color}.''
\end{quote}
Consider now the ASP program $P$ that consists only of the constraint 
(\ref{eq:colrule2}). This program has a  unique answer set $\emptyset$, 
the empty believed literal set. Let $B$ be any (informal) belief state that 
satisfies (*). In order for our proposed sentence (*) to be a correct 
informal semantics for (\ref{eq:colrule2}), this belief state $B$ must 
abstract to the unique answer set $\emptyset$ of $P$. If so, then in $B$ 
it is not known that $a$ is colored (because 
$Colored(a) \not\in \emptyset$). It follows that $B$ must satisfy the 
property ``\emph{it is not the case that a is a node}.'' 
In other words, $B$ cannot consider as possible any state of affairs in 
which $a$ is a node. Or, to put it in even simpler terms, the belief state 
$B$ must contain the belief that $a$ is not a node. However, since $B$ 
abstracts to $\emptyset$, it must then be the case that $\neg Node(a) \in 
\emptyset$, which is clearly false. Therefore, our proposed informal 
semantics of (\ref{eq:colrule2}) is incorrect.  

It is not the purpose of this article to present an epistemic informal 
semantics for modern ASP languages. We bring up this issue to point out 
that doing so is often more problematic than one might expect. This is 
not only the case for constraints, but also for choice rules, whose 
informal epistemic semantics has been the topic of lively debates in 
the ASP community\footnote{This is illustrated by
  the Texas Action Group discussion on the matter
  (\url{http://www.cs.utexas.edu/users/vl/tag/choice_discussion}).}. 
For the sake of completeness, we conclude this discussion by mentioning 
that a correct informal epistemic reading of ~(\ref{eq:colrule2}) is: 
\begin{quote}
{(**) ``\emph{for every $x$, it is not the case that \emph{the agent
knows that} $x$ is a node and the agent does not know that $x$ has a color}.''} 
\end{quote}
Writing constraints such as \eqref{eq:colrule2} is common practice among 
ASP programmers. At the same time, the fact that (*) is not a correct 
informal semantics for them while (**) is,  appears to be far from common 
knowledge. This illustrates the pitfalls of using the Gelfond and 
Lifschitz' epistemic informal semantics with the GDT programming methodology.
}



Our goal in
this paper is to develop an alternative informal semantics $\infsemT{\II}$ 
for answer set programming, which, unlike $\infsemGL{\II}$, is
\textcolor{black}{\emph{objective}}, that is, not epistemic,
and extends to new features of the language that
ASP now takes for granted. The rationale behind this effort is our conviction
that such an objective informal semantics will be better aligned with typical 
GDT programs.
Before we introduce it, we first review the GDT paradigm.

\section{Generate-Define-Test methodology}
\label{gdt}

The \emph{generate-define-test} (GDT) methodology \citep{lif02} was
proposed as a principled way to encode search problems in ASP.  Over the
years, it became the \emph{de facto} standard in the field.  
The GDT methodology yields programs that consist of three parts: 
\gen, \define and \test.

The role of \gen is to {\em generate the search space}. In modern
ASP languages this task is often accomplished by a set of \emph{choice
rules}:
\beq 
\{ A \} \rul B_1,\dots,B_n, \naf C_1,\dots,\naf C_m,
\eeq{eq:choice}
where $A$, $B_i$, and $C_j$ are atoms (possibly, non propositional). Intuitively, rule~\eqref{eq:choice}
states that the atom in the head can take \emph{any} of the values true and 
false, if the condition expressed by the body of the rule holds.
We call the predicate symbol of $A$ {\em generated}.  

The \define part is a set of definitions of {\em input} and {\em
  auxiliary} predicates. Each definition is encoded by a group of
\emph{normal rules} (i.e., rules of the form~\eqref{eq:rule} 
in which the expressions $A$, $B_i$ and $C_j$ are all atoms). In such a 
rule, the predicate symbol of the head $A$ is the one being defined. 
Input predicates 
are \emph{defined} by exhaustive enumeration of their extensions. 
That is, their definitions consist of sets of \emph{facts}, i.e., 
rules of form (\ref{eq:rule}) that have empty bodies and contain no
variables. For facts, the symbol $\leftarrow$ is often omitted from
the notation. Auxiliary predicates are \emph{defined} by sets of rules
\eqref{eq:rule} that specify these symbols by describing how to derive 
their extensions from the extensions of the generated and input symbols.

Finally, the \test part eliminates some generated candidate
answer sets. It consists of \emph{constraint rules}:  
\beq
\rul B_1, \dots, B_n, \naf C_1, \dots, \naf C_m,
\eeq{eq:constraint}
where $B_i$ and $C_j$ are atoms. Each constraint rule eliminates those 
candidate answer sets, in which all $B_i$ are true and all
$C_j$ are false. 

As such, the GDT methodology identifies different components of
programs by assigning them a particular operational task (e.g.,
``generate search space'', ``test candidate solution'') in the
computation of answer sets.  Thus, it is about the \emph{computation
  process} of answer sets and not about the \emph{meaning} of
expressions in ASP programs. In other words, the GDT
methodology does not specify an informal semantics for GDT
programs. 

We call finite sets of choice, normal, and constraint rules
\emph{\coreASP programs}. For these \coreASP programs, we will adopt the formal semantics proposed by \citeauthor{fer09}~\citeyearpar{fer09}. The main concepts of this semantics are briefly reviewed in the appendix; we omit the details, as they are not relevant for the purposes of this paper.

While syntactically simple, \coreASP programs are
expressive enough to support the GDT methodology. Moreover, they are a part 
of almost all ASP dialects, modulo minor syntactic differences. Thus, they 
form a convenient target language for our study of GDT programs. 
However, the GDT methodology is not restricted to \coreASP. 
Many applications rely on extensions such as 
aggregates and weight expressions, or use different means to implement
the \emph{generate} task (pairs of rules $P \leftarrow {\tt not}~P^*, 
P^*\leftarrow {\tt not}~P$; disjunctions $P\lor P^*$; or 
rules $P\leftarrow {\tt not\ not}~P$). The discussion we present here 
also applies to programs that contain such expressions. 
We touch on this subject 
in Section~\ref{SecDiscussion}.

To illustrate the GDT methodology, we present a \coreASP program that encodes the Hamiltonian cycle problem. 
\beq \small
\begin{array}{ll} 
\hbox{\gen}&   \{ \In(x,y) \} \rul \Edge(x,y). \\  \hline
\vspace{-6pt}\\
\hbox{\define}&\Vertex(V).\ \dots\ \Vertex(W).  \\
\cline{2-2}\\
\vspace{-16pt}\\
&   \Edge(V,V').\ \dots\ \Edge(W,W').\\ 
\cline{2-2}\\
\vspace{-16pt}\\
& T(x,y) \rul \In(x,y).\\
&   T(x,y) \rul T(x,z), T(z,y). \\ \hline
\vspace{-6pt}\\
\hbox{\test}& \rul \In(x,y), \In(x,z), y\neq z.\\ 
\cline{2-2}\\
\vspace{-16pt}\\
& \rul \In(x,z), \In(y,z), x\neq y. \\ 
\cline{2-2}\\
\vspace{-16pt}\\
& \rul \Vertex(x), \Vertex(y), \naf T(x,y).\\
\end{array}
\eeq{eq:hc-asp} The long horizontal lines indicate the partition of the
program into the {\gen}, {\define}, and {\test} parts
respectively. The short lines inside the {\define} part separate
groups of rules defining individual predicates. Inside the {\em test}
part, they separate individual constraints. Predicate
symbol $\In$ is a generated symbol, symbols $\Vertex$ and $\Edge$ are
input symbols, and $T$ is an auxiliary one.  The {\gen} part
specifies all subsets $\In$ of the $\Edge$ relation as candidate
solutions. The relations $\Vertex$ and $\Edge$ represent all
vertices and edges of the input graph and are defined by
enumeration. The relation $T$ represents the auxiliary concept of the
transitive closure of the relation $\In$. The {\test} part
``weeds out'' those candidate answer sets, whose $\In$ relation does 
not correspond to a Hamiltonian cycle. The auxiliary
relation $T$ is necessary to state the constraint that a Hamiltonian
cycle connects every pair of vertices.

As this example illustrates, a GDT program 
often has a rich internal structure. For instance, in the
example above 
\begin{itemize}
	\item \textcolor{black}{rules can be partitioned into three groups 
              \gen, \define and \test} 
	\item \define
	contains separate definitions for three predicates $\mathit{Node}$,
	$\Edge$ and $T$, and
	\item \test  consists of three independent constraints.
\end{itemize}
 The fact that the internal structure of programs remains implicit in the
standard ASP formalism motivates us to introduce the \NewASP
language in the next section that makes the structure of programs explicit.
In \NewASP, rules in \gen, \define and \test can be further split out in independent modules called \Gmodules, \Dmodules and \Tmodules. 

\section{The logic \NewASP}
\label{sec:newasp}

We now turn to our goal of developing an informal semantics for GDT
programs. To this end, motivated by the  GDT methodology, we  propose 
a logic \NewASP and develop for it an informal semantics. We design 
the logic \NewASP so that each \coreASP program can be cast as an 
\NewASP program without any essential changes. In this way, the informal 
semantics for \NewASP can be used for ASP programs \textcolor{black}{and, as 
we argue, becomes particularly effective in explaning the meaning of GDT 
programs.}

\subsection{Syntax}
\label{section:ASPFO}

As in \FO, expressions of \NewASP are built from predicate and function symbols  of some vocabulary $\Voc$.
Theories of \NewASP consist of three types of modules: \Gmodules, \Dmodules
and \Tmodules. 

A {\em choice rule} is an expression of the form: 
\beq
\forall \x\; (\{ P(\ttt) \} \rul \varphi),
\eeq{eq:gchoice}
where $\varphi$ is an FO formula, $P(\ttt)$ is an atom, and $\x$ includes 
all free variables appearing in the rule. We call the expression $\{P(\ttt)\}$
the \emph{head} of the rule and refer to $P$ as its \emph{head predicate}.
We call $\vph$ the \emph{body} of the rule.

\begin{definition}[\Gmodule] 
 A \emph{\Gmodule} is a finite set of choice rules
  with the same head predicate. Moreover, the head predicate may not
  appear in the body of the rules.
\end{definition}

While many modern ASP solvers allow recursive choice
rules, our concept of \Gmodules is more
restrictive.  This is in keeping with our view of
\Gmodules as modules that generate the search space from a given
problem instance.  To generate such a search space, recursion does not
seem to be required.

Recursion is allowed and, \cred{in fact, necessary} in the {\em define} part.  
A \emph{define rule} is an expression of the form
\beq\forall \x\;(P(\ttt) \rul \varphi),
\eeq{e:defrule}
where $\varphi$ is an FO
formula, $P(\ttt)$ is an atom, and $\x$ includes all free variables
appearing in the rule. The concepts of the \emph{head}, the \emph{head
predicate} and the \emph{body} of the rule are defined in an obvious way 
similarly as above. 

\begin{definition} [\Dmodule]
A \emph{\Dmodule} $\D$ is a pair $\struct{\exta,\Pi}$, where $\exta$ is
a finite set of predicates, called {\em defined predicates}, and $\Pi$ is a
finite set of define rules such that the head predicate of each rule belongs to $\exta$.  
\end{definition}

For a \Dmodule $\D$, we denote the set $\exta$ of its defined
predicate symbols by $\extp{\D}$. We write $\pars{\D}$ for the set of
all predicate and function symbols  in $\Pi$ other than the
defined predicates. We call $\pars{\D}$ the set of {\em parameter
  symbols} of $\D$.

For a set of define rules $\Pi$, by $\hd(\Pi)$ we denote the set of
all predicate symbols appearing in the heads of rules in $\Pi$. Note
that if $\Pi$ is the set of define rules of a \Dmodule $\D$, then
$\hd(\Pi)\subseteq \extp{\D}$. \textcolor{black}{If the inclusion is proper, 
the \Dmodules $\struct{\hd(\Pi),\Pi}$ and $\D$ are different; the latter 
makes all predicates in $\extp{\D} \setminus \hd(\Pi)$ universally false.}
In the following, we use $\Pi$ as a shorthand notation for a \Dmodule 
$\struct{\hd(\Pi),\Pi}$.

\begin{definition} [\Tmodule]
A \emph{\Tmodule} is an \FO sentence.
\end{definition}

While \Gmodules and \Dmodules are \emph{sets} of 
expressions, a \Tmodule is not. Since any finite set of 
\FO sentences $\varphi_1,\dots,\varphi_n$ can be equivalently 
represented by its conjunction $\varphi_1\land\dots\land\varphi_n$,
the restriction of \Tmodules to single formulas does not result
in any loss of generality. 

\begin{definition} \label{foaspth}
An \emph{\NewASP theory} is a finite set of \Gmodules, \Dmodules, and 
\Tmodules.
\end{definition}
 We say that an \NewASP module or theory $\Psi$ is \emph{over vocabulary $\Sigma$} if every non-logical symbol mentioned in $\Psi$ belongs to $\Sigma$. 
In case of a \Dmodule $\struct{\exta,\Pi}$, also symbols in $\exta$ 
should belong to $\Sigma$.

We say that a predicate is \emph{specified} by a \Gmodule or \Dmodule
if it is the head predicate of the \Gmodule or a defined predicate of the
\Dmodule.  Unless stated differently, we assume that no predicate is
specified by more than one module in an \NewASP theory as this suffices
to express GDT programs. However, the formal definitions of syntax and
semantics of \NewASP do not require 
this limitation.




\subsection{From Core ASP to \NewASP Informally}
\label{ssec:core2FO}

There is an obvious match between language constructs of \NewASP and 
those used in ASP to express \gen, \define and \test. Specifically, 
an ASP choice rule (\ref{eq:choice}) corresponds to an \NewASP\ 
choice rule
\[
\forall \x\;(\{ A \} \rul B_1\land \dots\land B_n\land  \neg C_1\land \dots
\land \neg C_m),
\]
a normal rule~(\ref{eq:rule}) corresponds to an \NewASP define rule
\[
\forall \x\;(A\rul B_1\wedge\cdots\wedge B_n\wedge \neg C_1\wedge\cdots\wedge
\neg C_m),
\]
and an ASP constraint~(\ref{eq:constraint}) corresponds to the \Tmodule 
given by the FO sentence
\[ \forall \x\; (\neg( B_1\land\dots\land B_n\land \neg
C_1\land\dots\land\neg C_m)),\] where in each case, $\x$ is the set of
variables occurring in the ASP expression (\ref{eq:choice}),
(\ref{eq:rule}) and (\ref{eq:constraint}), respectively. These
syntactical translations turn both the constraint operator
$\leftarrow$ and the negation-as-failure symbol {\tt not}
in~(\ref{eq:constraint}) into the negation symbol $\lnot$.

Consider the encoding~\eqref{eq:hc-asp} of the Hamiltonian cycle
problem. It can be embedded in \NewASP in several ways, including the following encoding  that
makes explicit the hidden structure of that program: 
\beq \small
\begin{array}{ll}
\hbox{\gen}\hspace{-0.05in}& \{\forall x \forall y (\{ In(x,y)\} \rul Edge(x,y)) \}\\
\hline
\vspace{-6pt}\\
\hbox{\define}& \{\Vertex(V)\rul\top, \dots, \Vertex(W)\rul\top\} \\
& \{Edge(V,V')\rul\top, \dots, Edge(W,W')\rul\top\}\\ 
& \defin{\forall x \forall y (T(x,y) \rul In(x,y))\\
\forall x \forall y \forall z (T(x,y) \rul  T(x,z)\land T(z,y))}\\\hline
\vspace{-6pt}\\
\hbox{\test}&
\forall x \forall y \forall z \neg(In(x,y)\land In(x,z)\land y\neq z)\\
 &  \forall x \forall y \forall z \neg(In(x,z)\land In(y,z)\land x\neq y)\\
 &  \forall x \forall y \neg(\Vertex(x)\land \Vertex(y) \land \neg T(x,y))
\end{array}
\eeq{eq:hc-asp-fo} 
Merging the three \Dmodules into one would yield another embedding, with less internal structure.

Any answer set program written in the base formalism of Section~\ref{gdt} has
a straightforward syntactic translation to an \NewASP theory: choice
rules are grouped according to their head predicate; normal rules are
grouped together in one \Dmodule, and each constraint can be
represented as a single \Tmodule (in each case, after rewritings
described above). In the case of a \Dmodule we can often reveal its
hidden structure by splitting it into several smaller \Dmodules (as we did in the theory (\ref{eq:hc-asp-fo})).
Section~\ref{sec:asp-rel} provides a detailed  formal account on the
relation between answer set programs and \NewASP theories.

\subsection{Semantics}\label{section:ASPFOsem}

We now introduce the formal semantics of \NewASP. As in FO and in a growing number of  ASP semantics, the semantics is based on the standard notion of $\Voc$-structures instead of literal sets. Using the terminology of logic programming and ASP, we call $\Voc$-structures also {\em $\Voc$-interpretations}.    

A crucial feature of the semantics of \NewASP is 
\emph{modularity}:
a structure/interpretation is a model of an {\NewASP} theory $T$ if it 
is a model of each of its modules. In other words, an {\NewASP} theory can 
be understood as a standard \emph{monotone conjunction} of its modules. 
\begin{definition}
  An interpretation $\I$ \emph{satisfies} (is a \emph{model} of) an
  \NewASP theory $T$, written $\I\models T$, if $\I$ satisfies 
(is a model of) each
  module in $T$.
\end{definition}


To complete the definition of the semantics of \NewASP 
theories, we now define the semantics of individual modules.

\paragraph{\Tmodules} 
The case of \Tmodules is straightforward. \Tmodules are \FO sentences
and we use the classical definition of satisfaction
(cf. Definition~\ref{deftrue}) to specify when a \Tmodule holds in a
structure.

\paragraph{\Gmodules}
The role of
choice rules in GDT programs is to ``open up'' some atoms~$P(\ttt)$
--- to allow them to take any of the values true and false.  We take
this idea as the basis of our formalization of the semantics of
\Gmodules.

\begin{definition}\label{def:gm}
An interpretation $\M$ is a model of a \Gmodule $\mathcal{G}$ with head 
predicate~$P$ if for each tuple $\bar{d}$ of elements in the domain of $\M$ such that 
$\M[\x:\dd]\models P(\x)$, there is a choice rule 
$\forall \y\; (\{ P(\ttt) \} \rul \varphi)$
in $\mathcal{G}$ and a tuple $\dd'$ so that 
$\M[\x:\dd,\y:\dd']\models \varphi$.
\end{definition}

\paragraph{\Dmodules}
We define the semantics of \Dmodules by adapting the stable semantics
of definitions introduced by \citet{pelov}. That semantics is based on
the three-valued immediate consequence operator. It is obtained as a
special case from the approximation fixpoint theory \citep{DeneckerMT00},
which defines stable and well-founded fixpoints for arbitrary lattice 
operators.  The semantics proposed by \citet{pelov} is a 
generalization of the original Gelfond-Lifschitz formal semantics. In 
particular, unlike the original semantics it is not restricted to Herbrand 
interpretations only. Our presentation follows that proposed by \citet{VennekensWMD07} and  developed further by \citet{DeneckerBV12}. 

\begin{definition}[Satisfaction by pairs of interpretations]
  \label{defdef} 
Let~$\varphi$ be an \FO formula, $\I$ and $\J$ interpretations
  of all free symbols in~$\vph$ (including free variables) such that $\I$ and $\J$ have the same 
  domain and assign the same values to all function symbols. We define
the relation $(\I,\J) \models \varphi$ by induction on the structure 
of~$\varphi$ (for simplicity, we consider only the connectives 
$\neg$ and $\lor$, and the existential quantifier $\exists$):
\begin{itemize}
\item[$-$] $(\I,\J) \models P(\ttt)$ if $\I \models P(\ttt)$;
\item[$-$] $(\I,\J) \models \neg \varphi$ if $(\J,\I) \not 
\models \varphi$;
\item[$-$] $(\I,\J)
\models \varphi \lor \psi$ if $(\I,\J) \models \varphi$ or
$(\I,\J)\models \psi$;
\item[$-$]  $(\I,\J) \models \exists x\ \psi$ if for some 
$d\in\dom{\I}$,
\mbox{\ }$(\I[x:d],\J[x:d])\models \psi$.
\end{itemize}
When $\vph$ is a sentence, we define $\varphi^{(\I,\J)}=\Tr$ 
if $(\I,\J)\models\varphi$ and $\varphi^{(\I,\J)}=\Fa$ otherwise.
\end{definition}
This is the standard satisfaction relation, except that positive
occurrences of atoms are interpreted in $\I$, and negative occurrences
in $\J$.
When $\I=\J$, 
this relation collapses to the standard satisfaction relation of \FO so that 
if $\vph$ is a sentence then $\varphi^\I=\varphi^{(\I,\I)}$.
\textcolor{black}{Also if $\I\leqt \I'$ and $\J'\leqt \J$, then
$\varphi^{(\I,\J)}\leqt\varphi^{(\I',\J')}$ (and so, in particular,
$\varphi^{(\cdot,\cdot)}$ is monotone in its first and antimonotone in
its second argument). These two properties imply that the satisfiability 
relation in Definition \ref{defdef} can be used to approximate the 
standard truth value, in the sense that if $\I_l \leqt \I \leqt \I_u$, 
then for each sentence $\varphi$, $\varphi^{(\I_l,\I_u)}
\leqt \varphi^\I\leqt \varphi^{(\I_u,\I_l)}$.} 

Essentially, this satisfaction relation represents Kleene's and
Belnap's three- and four-valued truth assignment functions
\citep{jsyml/Feferman84}. The connection is based on the bilattice
correspondence between the four truth values $\Fa,\Tr,\Un,\Inc$ and
pairs of lower and upper estimates $(\Fa,\Fa),
(\Tr,\Tr), (\Fa,\Tr)$, and $(\Tr,\Fa)$, respectively.  In this view, three- and
four-valued interpretations $\tilde{\I}$ correspond to pairs of
interpretations $(\I_l,\I_u)$ sharing the domain and the
interpretations of function symbols, and the
truth value $\varphi^{\tilde{\I}} \in \{\Fa,\Tr,\Un,\Inc\}$ corresponds to the pair
$(\varphi^{(\I_l,\I_u)},\varphi^{(\I_u,\I_l)})$.

\begin{definition}
\label{def:sat-mt2}
A pair of interpretations $(\I,\J)$ sharing the domain and the 
interpretations of function symbols \emph{satisfies} a define rule 
$\forall\x\; (P(\ttt) \rul \varphi)$ if for each tuple $\dd$ of domain elements, if $(\I[\x:\dd],\J[\x:\dd])\models \varphi$ then $\I[\x:\dd]\models P(\ttt)$. 
\end{definition}

In the context of Herbrand interpretations, the two 
definitions reflect the way Gelfond and Lifschitz used the reduct to define 
stable models of normal logic programs. Let us recall that an extended logic 
program is \emph{normal} if it consists of normal program rules only, and
let us consider a \Dmodule that corresponds to a normal program $\Pi$. Let 
$\M$ be an Herbrand interpretation. For a body $\varphi$ of a rule in 
$\grnd(\Pi)$ (the ground instantiation of $\Pi$), we write $\varphi^\M$ for
its \emph{reduced} form obtained by replacing the negative literals in 
$\varphi$ by their evaluation in $\M$. Clearly, for every rule $P\rul \varphi$
in $\grnd(\Pi)$, an Herbrand interpretation $\I$ satisfies the reduced
body $\varphi^\M$ if and only if $(\I,\M) \models \varphi$. Since the reduct
$\grnd(\Pi)^\M$ can essentially be viewed as obtained from $\grnd(\Pi)$ by
replacing the body of each rule by its reduced form, we have the following
property explaining the connection of Definitions \ref{defdef} and 
\ref{def:sat-mt2} to the
concept of a stable model of a normal program.

\begin{proposition}[\cite{DeneckerBV12}] \label{prop:simul}
For a normal program $\Pi$ and Herbrand interpretations $\I$ and $\M$, 
the interpretation $\I$ is a model of the 
reduct~$\grnd(\Pi)^\M$ if
and only if the pair of interpretations $(\I,\M)$ satisfies all rules of 
$\Pi$ (viewed as define rules). {Further, $\M$ is a stable 
model of $\Pi$
if and only if $\M$ is the $\leq_t$-least Herbrand interpretation $\I$
such that $(\I,\M)$ satisfies all rules of $\Pi$ (viewed as define rules).}
\end{proposition} 

In the setting of \Dmodules we must account for the parameters (input symbols)
that may appear in the rules. For two structures $\I$ and $\J$ that have 
the same domain and interpret disjoint vocabularies, let $\I\circ\J$ denote 
the structure that 
\begin{enumerate}
\item interprets the union of the vocabularies of $\I$ and $\J$, 
\item has the same domain as $\I$ and $\J$, and 
\item coincides with $\I$ and $\J$ on their respective vocabularies. 
\end{enumerate}
Following Gelfond and Lifschitz, to define a parameterized version of 
the stable-model semantics, we perform minimization with respect to
the truth order $\leq_t$.

\begin{definition}[Parameterized stable-model semantics] \label{defsemD}
  For a \Dmodule~$\D$, an interpretation $\M$ of $\extp{\D}$ is a {\em
    stable model} of $\D$ relative to an interpretation $\I_p$ of
  $\pars{\D}$ if $\M$ is the $\leq_t$-least among all interpretations $\I$
  of $\extp{\D}$ such that $\I$ has the same domain as $\I_p$
and $(\I_p\circ\I,\I_p\circ\M)$ satisfies all rules of $\D$.%
\footnote{It is a simple
consequence of the monotonicity of $(\I,\J)\models\varphi$ in $\I$ that 
this $\leq_t$-least interpretation always exists \citep{VennekensWMD07}.} 
\end{definition}

This parameterized stable-model semantics of \Dmodules extends the 
original stable-model semantics of normal logic programs
in three ways: 
\begin{enumerate}
	\item
	it is parameterized, that is, it builds
	stable models on top of a given interpretation of the parameter
	symbols; 
	\item it handles FO bodies; and 
	\item it works for arbitrary (also 
	non-Herbrand) interpretations.
\end{enumerate} 
It shares these properties with other
recent generalizations of the original ASP formalism, such as
\citet{pea05}, \citet{lee08}, \citet{fer09}, \citet{zhang10} \citet{zhou11} and \citet{asuncion12}. 

The parameterized stable-model semantics  turns a \Dmodule $\D$ into a {\em 
non-determi\-nistic} function from interpretations of $\pars{\D}$ 
to interpretations of $\extp{\D}$. An interpretation $\I$ \emph{satisfies}
$\D$ if it \emph{agrees} with the function defined by $\D$, i.e., if its interpretation of $\extp{\D}$ is one of the possible images under this 
function of its interpretation of ${\pars{\D}}$.

\begin{definition}[The semantics of \Dmodules] \label{defstable}
An interpretation $\I$ is a \emph{model} of a \Dmodule $\D$, written
$\I\models \D$, if $\I|_{\extp{\D}}$ is a stable model of $\D$ relative to 
$\I|_{\pars{\D}}$. 
\end{definition}

We stress that we use the term \emph{model} here to distinguish the
present concept from 
the stable model relative to an interpretation (Definition \ref{defsemD}).

\begin{example}
\label{ex:1}
Let us consider a \Dmodule:
\[
\D =\left\{ \begin{array}{l}
            \forall x\; (p(x) \rul \neg q(x))\\
            \forall x\; (q(x) \rul \forall y\; r(y,x)).
            \end{array}
    \right\}
\]
There are no function symbols here, $r$ is the only parameter 
of $\D$, and $p$ and $q$ are the defined symbols. Each interpretation 
$\I_p$ of the parameter $r$ determines the corresponding set of 
stable models relative to $\I_p$. Each such stable model is an interpretation 
of the defined symbols $p$ and $q$. Let us investigate how these stable
models are to be obtained. 

The class of candidates for a stable model relative to a given $\I_p$ 
consists of interpretations $\M$ of the defined symbols $p, q$ that 
have the same domain as $\I_p$. For each such $\M$, $\I_p\circ\M$ is 
an interpretation 
of \emph{all} symbols occurring in $\Pi$, that matches $\I_p$ on the 
parameters and $\M$ on the defined symbols. Let us fix one such $\M$ and
consider the set of all interpretations $\I$ such that 
$(\I_p\circ\I, \I_p\circ\M)$ satisfies this rule set. In the evaluation 
of the first rule, $q$ occurs negatively and so, it is evaluated with
respect to $\I_p\circ\M$. Moreover, since $q$ is a defined predicate, 
it is evaluated in $\M$. For the second rule, the parameter $r$ is 
evaluated in $\I_p$. Thus, the set of all interpretations $\I$ such that 
$(\I_p\circ\I, \I_p\circ\M)$ satisfies $\D$ contains each 
interpretation $\I$ such that
\begin{itemize}
\item[$-$] $q^\I \supseteq \{d\in dom(\I_p) | \forall d'\in dom(\I_p) : (d',d)\in r^{\I_p}\}$.
\item[$-$]  $p^\I \supseteq dom(\I_p)\setminus q^{\M}$
\end{itemize}
According to the definition, $\M$ is a stable model of $\D$ relative to 
$\I_p$ if it is the smallest interpretation of $p$ and $q$ to satisfy 
this condition. This is the case precisely if neither of these two set 
inclusions is strict, that is, if 
\begin{itemize}
\item[$-$]  $q^\M=\{d\in dom(\I_p) | \forall d'\in dom(\I_p) : (d',d)\in r^{\I_p}\}$
\item[$-$] $p^\M=dom(\I_p)\setminus q^{\M}=\{d\in dom(\I_p) | \exists d'\in dom(\I_p) : (d',d)\not\in r^{\I_p}\}$.
\end{itemize}
This shows that each interpretation $\I_p$ of $r$ determines a unique stable 
model of $\D$. 

Applying Definition \ref{defstable} to this example, we see that an 
interpretation $\I$ of the vocabulary $\Sigma=\{r,p,q\}$ is a model of 
the \Dmodule $\D$ if $\M=\I|_{\{p,q\}}$ satisfies the equations 
on $p$ and $q$ obtained from the equations for $p^{\M}$ and $q^{\M}$ by 
substituting $\M|_{\{r\}}$ for $\I_p$. 
\end{example}




The language design of \NewASP was guided by our aim to develop an
informal semantics for the most basic expressions and connectives of
ASP. It is straightforward to extend the language \NewASP with
additional types of modules and language expressions.
Section~\ref{section:herb} introduces one new module called an
\emph{Herbrand module}. 
In Section~\ref{SecDiscussion} we discuss other possible
extensions with aggregates, weight expressions and disjunction in the
head.

\ignore{
Other important extensions of ASP languages include aggregates (or
weight expressions), disjunction in the head, and strong negation.
Extending \Tmodules and \Gmodules with aggregates is
straightforward. They can also be allowed in the bodies of rules of
\Dmodules. An extension of the parameterized stable semantics for
aggregates was developed by Pelov et al.~\shortcite{pelov}. To handle
define rules with disjunctions in the heads of rules one can adjust
Definition~\ref{defsemD} by requiring that a stable model $\M$ be a
{\em minimal} rather than the \emph{least} interpretation satisfying
the conditions given there.  Finally, strong negation $\sim$ can be
``translated away'' 
by means of additional predicates. In Section \ref{SecDiscussion} we discuss these extensions from the point of view of informal semantics.}

\subsection{Herbrand Modules}\label{section:herb}
In applications where the domain of all relevant objects is known, it
is common to design a vocabulary $\Voc$ such that each domain element is
denoted by exactly one ground term. In such a case, considering Herbrand
interpretations is sufficient. ASP is tailored to
such applications.  Consequently, it typically restricts its semantics
to Herbrand interpretations only. The logic \NewASP  is an open
domain logic with uninterpreted function symbols. We now introduce an
Herbrand module into the language of \NewASP. Its role is to express the
proposition that the domain of discourse is the Herbrand universe. It
allows us to restrict the semantics of \NewASP to Herbrand
interpretations and will facilitate the formal embedding of (standard)
ASP into \NewASP.

\begin{definition}
An \emph{Herbrand} module over a set $\sigma$ of function symbols is the 
expression~$\Her{\sigma}$. We say that $\M$ is a {\em model} of $\Her{\sigma}$, denoted  $\M\models\Her{\sigma}$, if $\dom{\M}$ 
is the set of 
terms that can be built from $\sigma$, and if
for each such term $t$, $t^\M=t$.
\end{definition}
If $\Sigma_F$ is the set of all function symbols of $\Sigma$, then the models of the Herbrand module $\Her{\Sigma_F}$ are precisely the Herbrand interpretations of $\Sigma$. 

We now extend the definition of an \emph{\NewASP theory} as follows: an 
\emph{\NewASP theory} is a finite set of \Gmodules, \Dmodules, 
\Tmodules, and Herbrand modules.


An Herbrand module $\Her{\sigma}$ in \NewASP can be seen as a
shorthand for the combination of  the domain closure axiom
$\mathit{DCA}(\sigma)$ and the \FO unique name axioms
$\mathit{UNA}(\sigma)$. A combination 
$\mathit{DCA}(\sigma)$ and $\mathit{UNA}(\sigma)$  can also be expressed in \NewASP by means of D-
and T-modules.\footnote{It is a well-known consequence of the
  compactness theorem for \FO that $\mathit{DCA}(\sigma)$ cannot be
  represented in \FO if $\sigma$ contains at least one constant and
  one function symbol of arity $\geq 1$.} Denecker~(\citeyear{den00})
illustrated how the logic FO(ID) captures this combination. The same
method is applicable in \NewASP. The idea is to introduce a new
predicate symbol $U/1$ and then the \Dmodule:
\[
\defin{\dots\\
  \forall x_1 \dots \forall x_n (U(f_j(x_1,\dots,x_n)) \rul U(x_1)\land\dots\land U(x_n))\\
  \dots }\] that has one such rule as above for every function symbol
$f_j/n$ in $\sigma$. In the case of a constant symbol $C$, the rule
reduces to $U(C)\rul \top$.  This \Dmodule defines $U$ as the set of
all ground terms of $\sigma$. The following \Tmodule is added to
express that there are no other objects in the domain:
\[\forall x\ U(x).\]
Combining the above \Dmodule and \Tmodule with FO axioms $\mathit{UNA}(\sigma)$ yields an \NewASP theory whose models are (isomorphic to) the structures
satisfying $\Her{\sigma}$. Thus Herbrand modules are redundant in
\NewASP and serve only as useful and intuitive abbreviations.

\subsection{Formal Relation to ASP}\label{sec:asp-rel}

We now show that \NewASP is a conservative extension of the \coreASP language 
so that we can formally relate \coreASP programs and ASP-FO theories. 
For a \coreASP program $\Pi$, by $\wh{\Pi}$ we denote the collection of
rules obtained by rewriting the rules in $\Pi$ in the syntax of \NewASP as
illustrated in Section~\ref{ssec:core2FO}. Further, for a vocabulary $\Sigma$ 
we write $\Sigma_P$ and $\Sigma_F$ for the sets of predicate and function
symbols in~$\Sigma$, respectively.

What we are looking for is a connection
between an ASP program and an \NewASP theory that in the case of GDT programs
makes their implicit structure explicit. To establish such a connection, we
use the results by \citet{fer09a} on ``splitting''. Splitting is a common 
method for uncovering the structure of programs for their further analyses. 
For instance, 
\citeauthor{erdogan04}~\citeyearpar{erdogan04} use it to prove 
correctness of GDT programs.
  
Let $\Pi$ be a \coreASP program of the syntactic form considered
in Section~\ref{gdt} with rules of the form (\ref{eq:choice}),
(\ref{eq:rule}) and (\ref{eq:constraint}).  We define the {\em positive
predicate dependency graph}
of~$\Pi$ as the directed graph that has all predicates of $\Pi$ as its
vertices and that has an edge from a predicate $P$ to a predicate $Q$
whenever $P$ appears in the head of a rule that has a
\emph{non-negated} $Q$ in its body. We say that $P$ \emph{positively
depends} on $Q$ if there is a \textcolor{black}{non-zero length} path from $P$ 
to $Q$ in this graph.

\begin{definition}
  A partition $\{\Pi_1,\dots,\Pi_n\}$ of $\Pi$ 
  is a \emph{splitting} of $\Pi$ if: 
\begin{itemize}
\item[$-$] each $\Pi_i$ that contains a constraint is a singleton; 
\item[$-$] for each predicate $P$, all rules with $P$ in the head belong  to the same $\Pi_i$; and 
\item[$-$] if two predicates positively depend on each other, their rules 
belong to the same module $\Pi_i$.
\end{itemize}
\end{definition}

Splitting tends to decompose a program in components that can be
understood independently. For instance, the horizontal
lines in the Hamiltonian cycle program~\eqref{eq:hc-asp} identify a
splitting in  which each component has a simple and natural
understanding.

\begin{definition}
  A splitting $\{\Pi_1,\dots,\Pi_n\}$ of $\Pi$ is \emph{proper} if no
  module $\Pi_i$ contains  choice rules for two different predicates,
  or both a normal rule and a choice rule. In addition, no head predicate of a choice module may positively depend on itself.
\end{definition}

The following proposition is straightforward. We use here a simplified 
notation, where $\wh{\Pi}$ denotes the \Dmodule 
$\struct{\hd(\wh{\Pi}),\wh{\Pi}}$.
\begin{proposition}
If $\Pi$ has a proper splitting $\{\Pi_1,\dots,\Pi_n\}$ then
$\{\wh{\Pi}_1,\dots,\wh{\Pi}_n\}$ is a well-formed \NewASP theory.
\end{proposition}
For instance, the splitting of the Hamiltonian cycle
program~\eqref{eq:hc-asp} identified by the horizontal lines is a proper splitting, and its translation
is the \NewASP theory~(\ref{eq:hc-asp-fo}).

Typically, a program $\Pi$ with a proper splitting
$\{\Pi_1,\dots,\Pi_n\}$ is equivalent to the corresponding \NewASP
theory augmented with the Herbrand module. However, this is not the case for programs that contain predicates which do \cred{not} appear in the head of any rule. In \coreASP, such predicates are universally false. To obtain the same effect in \NewASP, we add an additional ``empty'' D-module $(\mathit{Def},\{\})$, where $\mathit{Def}$ is the set of these predicates.

\begin{theorem} \label{theoSPLIT}
  Let $\Pi$ be a \coreASP program over a finite vocabulary $\Voc$
  with a proper splitting $\{\Pi_1,\dots,\Pi_n\}$. Then an
  interpretation $\M$ is an answer set of $\Pi$ if and only if
  $\M$ is a model of the \NewASP theory $\{\Her{\Voc_F},\wh{\Pi}_1,\dots,\wh{\Pi}_n,(\mathit{Def},\{\})\}$, where $\mathit{Def} = \Voc_P\setminus\hd(\Pi)$.
\end{theorem}
This theorem essentially follows from the splitting result of \cite{fer09}. 
We present an argument in the \cred{appendix}, as it requires technical 
notation and concepts not related to the main topic of the paper.

Theorem~\ref{theoSPLIT} implies that answer sets of the GDT 
program~\eqref{eq:hc-asp} coincide with Herbrand models of the \NewASP
theory~\eqref{eq:hc-asp-fo}. 
More generally, Theorem~\ref{theoSPLIT} states that properly
splittable ASP programs can be embedded in \NewASP while preserving
their implicit internal structure. 
It therefore implies that such a program can be understood as the
\emph{monotone conjunction} of the \NewASP modules of which it
consists. Thus, even though ASP is a nonmonotonic logic, its 
nonmonotonicity is restricted to individual
\NewASP modules of which it consists.
In this way, the theorem paves the way towards an informal semantics of GDT
programs by reducing the problem to finding the informal semantics of
their generate, define and test modules.

\section{Informal semantics of \NewASP}\label{sec:informal}

In this section, we  develop the theory of informal semantics
$\infsemT{\II}$ for \NewASP.  We use the Hamiltonian cycle \NewASP theory~\eqref{eq:hc-asp-fo} as a  test case. Through the
model-preserving embedding of Theorem~\ref{theoSPLIT}, this informal
semantics applies  to the original GDT program~\eqref{eq:hc-asp}.  

The Hamiltonian cycle theory expresses that a graph $\In$ is a Hamiltonian 
cycle of graph $\Edge$: a linear cyclic subgraph of $Edge$ that includes all vertices. 
The intended interpretation $\II$ of the predicate symbols of this theory can be specified as follows: 
\textcolor{black}{
\begin{itemize}
\item {$\II(\Vertex)$: ``$\#_1$ is a vertex''}  
\item {$\II(\Edge)$:  ``there is an edge from $\#_1$ to $\#_2$  in graph \emph{Edge}''}
\item $\II(\In)$: ``there is an edge  from $\#_1$ to $\#_2$  in graph \emph{In}'' and 
\item $\II(T)$: ``$\#_2$ is reachable from $\#_1$ in graph \emph{In}''. 
\end{itemize}}

{In addition to the above predicate symbols, the vocabulary of this theory 
also contains a number of constant symbols $v,w,\ldots$. These are intended to represent the nodes of the graph. We therefore also add the Herbrand module $\Her{\Sigma_F}$ to the theory.}

\subsection*{The composition operator of \NewASP theories}

Formally, a structure $\I$ is a model of an \NewASP theory if it is a
model of each of its modules. In our Tarskian perspective, this means
that a world is possible according to a theory if it is possible
according to each of its modules. 
{Thus, as for FO, the composition operator 
that describes how the meaning of a theory depends on the meaning of its elements 
is simply the standard {\em conjunction}}: if 
an \NewASP theory
$T$ consists of modules $\Psi_1,\dots,\Psi_n$, then $\infsemT{\II}{\formulas}{T}$
is the conjunction of the statements $\infsemT{\II}{\formulas}{\Psi_i},  \ldots$$,
\infsemT{\II}{\formulas}{\Psi_n}$.  Therefore, adding a new module to
an \NewASP theory is a monotone operation, in the same way that adding
an additional formula to an \FO theory is.

Theorem~\ref{theoSPLIT} shows that a GDT program (to be precise, a core ASP 
program with a proper splitting) 
can be viewed as a \emph{monotone} conjunction of its
components. Thus, the nonmonotonicity of an ASP program in
the GDT style is confined to individual components. Indeed, we will
see that the informal composition operators that construct \Gmodules and
\Dmodules from individual rules are not monotone. That is, the meaning of 
\Gmodules
and \Dmodules cannot be understood as a simple conjunction of the meanings
of their rules.

\ignore{\marc{I dont think we say this this, but it is a fact that in the epistemic viewpoint, taking the intersection of the class of belief states of different components of a theory is NOT the conjunction operator. Rather, it is something like selecting the belief states that different agents share. There is no increase of knowledge, contrary to in conjunction. }}

\subsection*{Informal semantics of T-modules (FO sentences)}

Formally, a T-module $T$ consists of FO sentences under their classical
semantics. Therefore, we set $\infsemT{\II}{}{T} =
\infsemFO{\II}{}{T}$. In the case of theory~(\ref{eq:hc-asp-fo}), this 
yields the following readings for its T-modules. The T-module
$$\forall x \forall y \forall z \neg(In(x,y)\land In(x,z)\land y\neq z)$$
states that 
\textcolor{black}{\em for all $x$, for all $y$ and for all $z$, it is not 
the case that there are edges from $x$ to $y$ and from $x$ to $z$ in graph
$\In$  and that $y$ and $z$ 
are not the same.}
{This can be restated as: {\em each domain element is the
start of at most one edge in the graph $\In$.}}

The T-module
$$\forall x \forall y \forall z \neg(In(x,z)\land In(y,z)\land x\neq y)$$ has 
a similar reading,
which can be equivalently stated as: 
{{\em each domain element is the end of 
most one edge in the graph $\In$.}}
The T-module $$\forall x \forall y \neg(\Vertex(x)\land \Vertex(y) \land 
\neg T(x,y))$$ says that \textcolor{black}{\em for
every $x$ and for every $y$, it is not the case that $x$ and $y$ are nodes, 
and $y$ is not reachable from $x$ in graph $In$.}
{This can be equivalently stated as: {\em every node is
reachable from every other node in graph $\In$.}}

The three propositions above are precisely the properties that the graph 
$\In$ should satisfy to be a Hamiltonian cycle of the graph $Edge$. 
{They therefore represent precisely 
what the ASP programmer intended to encode.} 

\paragraph{Informal semantics of G-modules (choice rules)}

Choice rules in ASP are often explained in a computational way, as
``generators of the search space.''  In this section, we develop a 
declarative interpretation for choice rules in \NewASP. 

We start by rewriting G-modules using a process similar to {\em predicate completion}~\citep{Clark78}.
First, every choice rule~\eqref{eq:gchoice} in a \Gmodule 
is rewritten as 
\[
\forall \y\; (\{ P(\y) \} \rul \exists \x(\y=\ttt\land \varphi)).
\]
Second, all resulting choice rules, say, 
\[
\forall \x\; (\{ P(\x) \} \rul \varphi_i), \tag*{{\text{for }$i=1,\ldots,n$}}
\]
are combined into a single one:
\beq
\forall \x\; (\{ P(\x) \} \rul \varphi_1\lor \dots \lor
\varphi_n).
\eeq{eq:gcombined}
\ignore{The theorem below states that some formula exists with some property P. It seems that to claim that the rewriting above is validated the theorem below should state that this particular rewriting has property P.
	\begin{theorem}
		\label{prop:ch-ra}
		For every finite \Gmodule $\Gc$ with head predicate $P$, 
		there is an FO formula $\vph$ such that $\Gc$ is equivalent to 
		the \Gmodule $\Gc' = \{\forall \x\; (\{ P(\x) \} \rul \varphi)\}$,
		that is, for every interpretation $\M$, $\M\models \Gc$ if and only 
		if $\M\models \Gc'$.
	\end{theorem}
	Proof: Indeed, consider $\vph=\varphi_1\lor \dots \lor \varphi_n$ as in~\eqref{eq:gcombined}.
	It is easy to see that  $\Gc$ and  $\Gc'$ have the same models. 
	\mirek{This proof does not add anything really above what was already said. It is either not needed or has to be extended to a formal argument.}
	\phantom{a}\hfill QED
}
We denote the result of this rewriting of a \Gmodule $\mathcal{G}$ to a 
singleton \Gmodule by $S(\mathcal{G})$. It is evident that the rewriting
preserves models.

\begin{theorem}
	\label{prop:ch-ra}		
	Every \Gmodule $\Gc$ 
	 is equivalent to the singleton \Gmodule  $S(\Gc)$. 
\end{theorem}

This result is important because singleton \Gmodules have a simple 
representation as FO sentences.

\begin{theorem} 
	\label{prop:ch-r}
	An interpretation $\M$ satisfies a singleton \Gmodule $\{
	\forall\x\; (\{ P(\x) \} \rul \varphi) \}$ if and only if $\M$ satisfies
	FO sentence $\forall \x\;(P(\x) \mim \varphi)$.
\end{theorem} 
Proof. By Definition \ref{def:gm},
$\M$ satisfies  $\{
\forall\x\; (\{ P(\x) \} \rul \varphi) \}$ if and only if for
each variable assignment $\theta$
such that $\M,\theta\models P(\x)$ it holds that 
$\M,\theta \models \varphi$. This is precisely the condition for 
$\M$ to satisfy $\forall \x\;(P(\x) \mim \varphi)$. \phantom{a}\hfill QED


\smallskip
For instance, the singleton \Gmodule
\[
\{\forall x \forall y (\{ In(x,y)\} \rul Edge(x,y)) \}
\]
of the \NewASP theory~\eqref{eq:hc-asp-fo} corresponds to the \FO sentence
\beq
\forall x \forall y (\mathit{In}(x,y) \mim Edge(x,y)).
\eeq{eq:folgen}

\cred{We call the result of first rewriting a \Gmodule $\mathcal{G}$ to a 
	singleton \Gmodule and then translating the latter to FO the {\em completion} 
	of $\mathcal{G}$. We denote it by $Gcompl(\mathcal{G})$.}
The following consequence of Theorem \ref{prop:ch-ra} provides
the key property of $Gcompl(\mathcal{G})$.

\begin{cor}\label{cor:gmodulesFO}
	Every 
	\Gmodule $\mathcal{G}$ is equivalent to the  FO sentence $Gcompl(\mathcal{G})$.
\end{cor}

This corollary demonstrates that 
\Gmodules 
can be simulated by \Tmodules. It  follows that  \NewASP 
theories can be seen as consisting of FO sentences and D-modules. 

A \Gmodule $\mathcal{G}$ for a predicate $P$ is a set of choice rules
\beq
\{\;\{ P(\ttt_1)\}\rul \varphi_1,\ \ \ldots,\ \  
\{P(\ttt_n)\}\rul \varphi_n\;\}. 
\eeq{eq:gmoduleforp}

By Corollary~\ref{cor:gmodulesFO}, such $\mathcal{G}$ is equivalent to
\cred{$Gcompl(\mathcal{G})$:}\ignore{$\Phi_\mathcal{G}$:}
\[
\forall \y (P(\y)\;\Rightarrow\;\exists \x_1 (\y=\ttt_1 \land \varphi_1) \lor 
\cdots \lor\exists \x_n (\y=\ttt_n \land \varphi_n)).
\]
Thus, given some intended interpretation $\II$ for a vocabulary $\Sigma$,
and a \Gmodule $\mathcal{G}$ over $\Sigma$ of the form~\eqref{eq:gmoduleforp},
the informal semantics $\infsemT{\II}{}{\mathcal{G}}$ must be equivalent 
to the informal semantics \cred{$\infsemFO{\II}{}{Gcompl(\mathcal{G})}$}. With
this in mind, we define $\infsemT{\II}{}{\mathcal{G}}$ by restating 
$\infsemFO{\II}{}{Gcompl(\mathcal{G})}$ as follows:  
\begin{quote}
\emph{In general, for each $\vec{x}$, $P^\II[\vec{x}]$ is false. However, there are exceptions as expressed by the following rules:
\begin{itemize}
\item[$-$] If \emph{{$\infsemFO{\II}{}{\varphi_1}$}}, then it might be that
\emph{{$\infsemFO{\II}{}{P(\ttt_1)}$}}.\\ \dots
\item[$-$]  If \emph{{$\infsemFO{\II}{}{\varphi_n}$}}, then it might be that
\emph{{$\infsemFO{\II}{}{P(\ttt_n)}$}}.
\item[$-$] There are no other exceptions.
\end{itemize}}
\end{quote}

This definition implicitly specifies the meaning of the logical connectives 
occurring in choice rule bodies, the rule operator~$\rul$, and the composition 
operator that forms a \Gmodule out of its rules. We now make this meaning
explicit.

First, in the translation of \Gmodules to \FO sentences, choice rule
bodies $\varphi_i$ are treated as ``black boxes,'' that are simply
copied and pasted into \FO expressions.  This shows that choice rule
bodies in \NewASP not only look like, but in fact \emph{are} \FO
expressions, with all \FO logical symbols retaining their
informal semantics. In particular, this illustrates that the
negation operator in the bodies of choice rules of an \NewASP\
\Gmodule is just classical negation. 



Second, to explicate the informal semantics of the rule operator and the
composition operator of \Gmodules, 
%
we note that the informal reading $\infsemT{\II}{}{\mathcal{G}}$ 
interprets a \Gmodule as a {\em local
closed world assumption} (LCWA) on predicate $P$ (\emph{local} refers to
the scope of the assumption, which is restricted to $P$), but provides 
an exception
mechanism to relax it. Each rule of a \Gmodule expresses an exception to
the LCWA and reinstalls uncertainty, the \emph{open world assumption}
(OWA), on the head atom. For instance, the \Gmodule
\[
\{\{\In(x,y)\} \rul \Edge(x,y)\}
\]
states that ``The Hamiltonian path $(\In)$ is empty \emph{except} that if $(x,y)$ is an edge of the graph $G$, then $(x,y)$ might belong to it.''  We note that this yields an
informal but precise linguistic reading of $\mathcal{G}$ and of rules
in $\mathcal{G}$, a reading that is indeed equivalent to the informal
semantics of $\mathcal{G}$'s translation into \FO which states that
$\In$ is a subgraph of $\Edge$. 

It can be seen in $\infsemT{\II}{}{\mathcal{G}}$ that the rule operator
in \Gmodules has  unusual
semantic properties.  Each rule of a \Gmodule is a conditional ``if
\dots then $P(\ttt)$ might be true''. Its ``conclusion'' removes
information (namely that $P(\ttt)$ is false) rather than adding
some. This is unlike any other \cred{conditional or expression in logic} that
we are aware of. 

Also the semantic properties of the composition operator of
\Gmodules are  unique.  The composition operator underlying
\Gmodules is neither conjunction nor disjunction. It is also 
not truth
functional and not monotone. Adding a rule to a module corresponds to
adding a disjunct to the corresponding \FO sentence. Hence, the
underlying composition operator is {\em anti-monotone}: the module
becomes {\em weaker} with each rule added. This agrees with the role
of a choice rule for expressing an {\em exception} to the LCWA on
$P$. The more rules there are, the more exceptions and hence, the
weaker the LCWA. 

To recap, \Gmodules are given a precise informal
semantics $\infsemT{\II}{}{\mathcal{G}}$ as a form of LCWA with an
exception mechanism to relax it. Logical connectives in the bodies of
choice rules retain their  classical  meaning. From a logical
point of view, the rule operator and the composition operator of
\Gmodules have uncommon semantical properties. Still,
$\infsemT{\II}{}{\mathcal{G}}$ identifies a natural
language conditional that explains formal choice rules in a declarative way. For example, when applied to the G-module of ASP-FO theory~(\ref{eq:hc-asp-fo}), it
yields a correct  reading of its \Gmodule.

\subsection*{Informal semantics of D-modules}


\cred{Humans use definitions to express abstractions of concepts they
encounter. These abstractions are necessary for us to understand the world 
in which we live and function, and to the ability to relay this understanding 
to others. We communicate these definitions in natural language; already as 
children, we are trained to compose, understand, and use them 
effectively.} Definitions also appear in the rigorous setting of
scientific discourse. In fact, they are the main building blocks of
formal science. In scientific and mathematical texts, definitions embody
the most precise and objective forms of human knowledge. While definitions
in a mathematical and scientific context are typically formulated with more 
precision than the definitions we use in everyday life, they are still 
{\em informal}, in the sense that they are not written in a formal language. 
We therefore refer to the unambiguous, precise natural language definitions of concepts
we find in everyday life or in science and mathematics as \emph{informal 
definitions}.

The stated goal of \Dmodules of an \NewASP theory (and the \define 
components of a GDT program) is to define concepts formally. We will 
now provide \Dmodules with an informal semantics matching precisely 
the formal one. The linguistic constructs used by humans to so effectively 
specify (informal) definitions are natural candidates for that task. 
{We therefore start by reviewing some of these natural language expressions.}

While there are no ``official'' linguistic rules on 
how to write an informal definition in a mathematical or scientific text, 
several conventions exist.  Simple definitions often take the form of ``if'' or
``if and only if''-statements.  More complex cases are inductive
(recursive) definitions, which are frequently represented as a set of
informal rules, possibly with an induction order. A good example is
Definition~\ref{deftrue}, where the satisfaction relation~$\models$
is defined over the subformula induction order. When written
according to these linguistic conventions, a definition has a
precise and objective meaning to us. 
Consider an
intended interpretation $\II$ for a vocabulary $\Sigma$,
a \Dmodule $\D = \langle \mathit{Def},\Pi\rangle$ in this vocabulary, with
$\mathit{Def}
= \{P_1,\dots,P_n\}$, and 
$\Pi =
\{\forall \x_1 (A_1\rul\varphi_1),\dots,\forall \x_m (A_m\rul\varphi_m)\}$. Assume
the following translation $\infsemT{\II}{}{\D}$ of $\D$ into natural
language.
\begin{quote} 
\emph{We define the relations $P_1^\II,\ldots,P_n^\II$ in terms of $\pars{\D}^\II$ by the following 
(simultaneous) induction:
\begin{itemize}
\item[$-$] \emph{$\infsemT{\II}{}{A_1}$}  if \emph{$\infsemT{\II}{}{\varphi_1}$}\\
\dots 
\item[$-$] \emph{$\infsemT{\II}{}{A_m}$} if \emph{$\infsemT{\II}{}{\varphi_m}$}
\item[$-$]  In no other cases,  $P_1^\II,\ldots,P_n^\II$ hold.
\end{itemize} }
\end{quote}
This last clause (``In no other cases ...'') is usually left implicit  if it is
clear that we are giving a definition. The question that we  address here is whether this is a suitable informal semantics for \Dmodules.

The above translation turns a \Dmodule $\D$ into a natural language
statement that follows the linguistic conventions used to
express inductive definitions. 
{If the \Dmodule is not recursive,} the phrase ``by the following simultaneous induction''
should be dropped; what remains then is a definition by
exhaustive enumeration, in which each rule represents one case.
This translation again makes use of the  natural
language connective ``if''.  As before, however, when this word is
encountered in the context of a case of an inductive definition, it
has a precise and unambiguous meaning which is clear to any mathematician. We
refer to this meaning as the ``definitional implication''.  We later
discuss how this conditional relates to material implication.

We now  test the stated informal semantics on the three \Dmodules of 
the Hamiltonian cycle theory~\eqref{eq:hc-asp-fo}. \textcolor{black}{The
first two modules correspond to non-recursive definitions by exhaustive 
enumeration of elements in the extensions of the $\Vertex$ and $\Edge$
relations, respectively.} The reading $\infsemT{\II}{}{\D}$ of
the remaining \Dmodule is as follows:
 \begin{quote} \emph{We define  $T$ in terms of the graph $\In$ by induction:
\begin{itemize}
\item[$-$] \textcolor{black}{for every $x$, for every $y$, $y$ is reachable 
from $x$ in the graph $\In$ if there is an edge from $x$ to $y$ 
graph in $\In$}
\item[$-$] \textcolor{black}{for every $x$, for every $y$, for every $z$,
$y$ is reachable from $x$ in the graph $\In$ if $z$
is reachable from $x$ in the graph $\In$ and $y$ is reachable from 
$z$ in the graph $\In$}
\end{itemize}}
\end{quote}
This is a standard monotone inductive definition of
the transitive closure of graph $In$, which is the intended
interpretation of $T/2$. 

Thus, we now have a proposal for a precise informal semantics
$\infsemT{\II}{}{\cdot}$ for \Dmodules, and, through the embedding result
of Theorem~\ref{theoSPLIT}, therefore also for \define components in GDT programs. \cred{The informal semantics we specified reflects
the role of these \define components in the GDT-methodology.}  
The
rest of the section is concerned with the following questions:
\begin{description}
\item[(a)] For which \Dmodules $\D$ is $\infsemT{\II}{}{\D}$ a sensible
informal definition of the relations in $\mathit{Def}(\D)$?
\item[(b)]\label{item.b} If $\infsemT{\II}{}{\D}$ is a sensible informal definition, are
  the relations defined by this informal definition indeed {\em the
    same} relations as produced by the parametrized stable semantics
  of the \Dmodule?
\item[(c)] What is the meaning of  the logical connectives in such \Dmodules and, through the embedding of GDT programs, in \define components of  GDT programs?
\end{description}

In the case of \FO, and therefore also of \Tmodules and \Gmodules,
the correspondence between the informal
semantics $\infsemFO{\II}{}{\varphi}$ and the formal semantics  of FO 
is made plausible by the great similarity between Table~\ref{fig:FOinf:form} and Definition \ref{deftrue}.
 In the case of \Dmodules, however,
the situation is  more complex. The reason for this is
that there is no obvious connection between the way (parametrized)
stable models are defined and the way we understand informal inductive
definitions.

To address the questions (a) - (c), we will borrow from the work on the logic 
FO(ID) \citep{den00,den08}. The logic FO(ID) was conceived as a conservative 
extension of FO with a formal \emph{definition} construct. Syntactically, 
FO(ID) corresponds to the fragment of \NewASP 
without \Gmodules. A \emph{definition} in the logic FO(ID) is a set of rules 
of exactly the same form as rules in \Dmodules. There is however a semantic 
difference: formal definitions in FO(ID) are interpreted under the two-valued 
parametrized \emph{well-founded} semantics rather than the stable semantics.

\ignore{
The three questions formulated above for \Dmodules of the logic \NewASP
arise also for FO(ID)'s definitions. They were investigated in detail by
Denecker and Vennekens (\citeyear{KR/DeneckerV14}), who compared the 
induction process(es) of a formal definition $\D$ (in the context of a 
parameter structure $\I$) with the well-founded model construction for 
$\D$. Let us call a definition $\D$ of FO(ID) (or a \Dmodule $\D$ of 
\NewASP) \emph{total} in a $\pars{\D}$-structure $\I$ if the parametrized
well-founded model of $\D$ in $\I$ is two-valued. \citet{KR/DeneckerV14} 
argued that the induction processes and the well-founded model construction 
for \emph{total} definitions converge to the same model. It is remarkable 
that the well-founded construction obtains this limit {\em without the
knowledge of the underlying induction order.} It recovers it from the 
structure of the rules. 
}

\textcolor{black}{
The three questions formulated above for \Dmodules of the logic \NewASP
arise also for FO(ID)'s definitions. They were investigated in detail by
Denecker and Vennekens (\citeyear{KR/DeneckerV14}). The view taken in that
study is that an  informal inductive or recursive definition  {\em defines} a set by specifying how to construct it: starting from the empty set, it proceeds by iterated rule application (possibly along some induction order) until the constructed set is saturated under rule application. 
Denecker and Vennekens formalized this \emph{induction process} for 
FO(ID) definitions $\D$ parameterized by a $\pars{\D}$-structure $\I$, 
and compared it with the well-founded model construction. They argued that 
if the well-founded model is two-valued, the  informal semantics $\infsemT{\II}{}{\D}$ of such a formal rule set is a sensible inductive definition and proved that all formal induction processes converge to the well-founded model. If the well-founded model is not two-valued, then the  informal semantics $\infsemT{\II}{}{\D}$ of such a formal rule set is not a sensible informal definition and the induction processes do not converge. This motivated them to call a definition $\D$ of FO(ID)  \emph{total} in a $\pars{\D}$-structure $\I$ if the parametrized well-founded model of $\D$ in $\I$ is two-valued. Their punch line is that 
{\em for definitions  $\D$ that are total in $\I$, the informal semantics $\infsemT{\II}{}{\D}$ presented above is a sensible informal definition and the relations that it  defines are given by the well-founded model of $\D$ extending $\I$}.}



We now observe that the logics FO(ID) and \NewASP are tightly related, not only syntactically, 
but also semantically. As long as we restrict attention to \Dmodules that 
have two-valued well-founded models, both logics are identical. 

\begin{theorem}[\citeauthor{pelov} (\citeyear{pelov})]
\label{pel07}
Let $\D$ be a formal definition of FO(ID) or a \Dmodule of \NewASP,
and let $\I$ be a $\pars{\D}$-structure. If $\D$ is
total in $\I$, then
the parameterized well-founded model of $\D$ in $\I$ is the unique
parametrized stable model of $\D$ that expands $\I$.
\end{theorem}

Thus, we obtain an answer to question (b) for 
the logic \NewASP. Provided that a \Dmodule $\D$ is total in $\I$ 
(a $\pars{\D}$-structure), $\infsemT{\II}{}{\D}$ is a correct and precise 
informal semantics for $\D$ {\em under the parametrized stable-model 
semantics}.

The totality condition on \Dmodules (or FO(ID) definitions) addresses 
the question (a) as it serves as a general semantic criterion for a sensible
definition. Broad classes of \Dmodules (FO(ID) definitions) are total in 
every context $\I$. Others are total only in some contexts $\I$. To provide 
some sense of scope, the classes of non-recursive, definite, stratified and
locally stratified logic programs have straightforward generalizations as
\Dmodules in \NewASP (definitions in FO(ID)). Non-recursive, definite, and 
stratified normal programs give rise to \Dmodules that are total in every 
interpretation of the parameter symbols \citep{den08} and can be thought 
of as formalizations of sensible definitions in every context $\I$. Locally 
stratified normal programs give rise to definitions that are total in
any Herbrand interpretation, and hence in the context of any theory
that contains $\Her{\sigma}$.

\yu{2. A question that
  rises. Is it possible to recast the arguments from  (Denecker and
  Vennekens 2014) to Definitions/D-modules under  stable models semantics of this
paper? This way we could construct a direct argument and instead of
reviewing  (Denecker and  Vennekens 2014) conclude by saying that the
nature of arguments of D modules under stable model semantics and D
modules under well-founded semantics is the same/closely related and
in fact follows the lines of (Denecker and  Vennekens 2014). 
My impression  is that this would make paper more self-contained and
add clarity. MARC: UNFORTUNATELY, THIS IS REALLY IMPOSSIBLE. THE ARGUMENTS IN THAT PAPER ARE FOR CONSTRUCTIONS IN 3-VALUED LOGIC WHICH ARE ALIEN TO STABLE SEMANTICS. 
}

\paragraph{Informal semantics of connectives in \Dmodules in \NewASP and FO(ID)}

We now address question (c).  Just as in the case of \Gmodules, the
informal semantics $\infsemT{\II}{}{\D}$ of \Dmodules implicitly
determines the informal semantics of the logical connectives occurring
in \Dmodule rule bodies and of the rule operator $\rul$.  Moreover, it also determines
the semantical composition operator that ``forms'' the meaning of a
\Dmodule from its rules.  The discussion below is restricted to
\Dmodules $\D$ for which $\infsemT{\II}{}{\D}$ is a sensible inductive
definition.

The translation $\infsemT{\II}{}{\D}$ treats a rule body $\varphi$ by
simply applying the standard informal semantics of FO to it.  All
logical symbols in rule bodies therefore retain their classical
meaning. In particular, the negation symbol in rule bodies of a
\Dmodule, and therefore, the negation as failure symbol in rule bodies
of GDT-programs is classical negation. After nearly 40 years of
controversy on the nature of negation as failure, this can be called a
surprising conclusion.

The rule operator $\rul$ in \Dmodules represents the sort of
conditional that is found in inductive definitions in mathematical
text. For example, we can phrase \[\infsemT{\II}{}{\forall x (Even(S(x)) \rul
  \neg Even(x))}\] as the conditional ``$n+1$ is even if $n$ is not
even''. In the context of a definition, such a rule sounds like a
material implication. However, while it indeed entails the material implication (i.e., its head must be true whenever its body is true),
 it is in fact 
much stronger than that (in particular, its head may not be arbitrarily true) and is not even a truth functional object.  In
particular, each rule is an ``instruction'' in the iterative
``recipe'' provided by an informal definition to construct the defined
relations.  This is the third kind of conditional that we encounter in this
paper.   In other studies of inductive definitions, this kind of conditional has also been called a \textcolor{black}{{\em production} \citep{martinlof1971}.}

The remaining question concerns the global informal composition
operator of \Dmodules. In mathematical text, this composition operator
and the modular nature of definitions surface most clearly when an
existing informal definition is extended with new cases. For instance,
the syntax of modal propositional logic may be derived from that of
propositional logic by a phrase such as: ``We extend the definition of
propositional logic with the additional case that if $\varphi$ is a
formula, then so is $K\varphi$''. In our terminology, this natural
language statement is invoking the informal composition operator of
inductive definitions to add an additional rule to an existing
definition. Such an additional rule has an impact on the construction
process specified by the definition, and therefore also on the
relation that is eventually constructed. After the addition, the
defined set of formulas becomes strictly larger since more formulas
can be constructed. However, the extension has a non-monotonic effect (in the sense used in the area of non-monotonic reasoning).  Indeed,
before the addition, the definition entailed for each propositional symbol
$p$ that $Kp$ was not a formula; after adding the rule, the definition
entails that $Kp$ is a formula. This is a revision and neither
monotone nor antimonotone.

We observe that from a logical perspective, the rule operator and the
global \Dmodule composition have very unusual properties and are truly
non-classical. In themselves, they are not  truth-functional. Yet, the 
\cred{definitions they form are ---}\ignore{meaning
composed of a set of rules forms a truth-functional object:} 
{a definition 
expresses a particular logical relation between parameter and defined
symbols that can be true or false in structures interpreting these
symbols.}  In summary, these non-standard features do not stop
human experts from understanding 
inductive definitions and the compositional nature of definitions, allowing 
them e.g., to  properly judge how an additional case changes the defined
concept. 

\subsection*{How much of ASP practice is covered by $\infsemT{\II}{}{\cdot}$? }  

The transformation to \NewASP in Theorem \ref{theoSPLIT} is equivalence 
preserving. 
Consequently, $\infsemT{\II}{}{\cdot}$ provides a precise
informal semantics that captures the content of splittable
ASP programs, under the condition that the resulting \Dmodules are total.  
Here, we assess how much of ASP practice is covered by these translations.

\ignore{
Recall that Theorem~\ref{thm:asp-a} translates a normal logic program $\Pi$
over $\Voc$ to the \NewASP theory
$\{\Her{\Voc_F},\struct{\Voc_P,\wh{\Pi}}\}$. Therefore, the function
$\infsemT{\II}{}$ provides a correct informal semantics if
$\struct{\Voc_P,\wh{\Pi}}$ is total in the context of Herbrand
structures. This is the case for non-recursive, definite, stratified,
locally stratified programs and any other normal program with a
two-valued well-founded Herbrand model. This covers many pure Prolog
and Datalog programs but almost none of the \textcolor{black}{early} GDT programs. The
reason is the presence of rules $\Aux \rul \neg \Aux, \dots$ to encode
\test components and rules $P \rul \neg P^*,\dots $ and $P^* \rul \neg
P, \dots$ to encode \gen components.
}

Theorem~\ref{theoSPLIT} applies to 
\coreASP programs $\Pi$ that have a proper splitting. 
Experience suggests that \define components in GDT programs
map frequently to classes of \Dmodules that are known to be total
(non-recursive, definite, stratified, locally stratified). For
example, in the Hamiltonian cycle program~\eqref{eq:hc-asp}, two
\Dmodules are non-recursive and one is negation-free.  In an
attempt to verify this on a broader scale, we  examined  benchmark programs of the 2013 ASP system competition \citep{asp4}.
In this experiment, we used an extended version of \NewASP that
supports weight constraints and aggregate expressions which occur in
many practical ASP programs. The reading $\infsemT{\II}{}$ and
Theorem~\ref{theoSPLIT} can be generalized for this formalism (see
the next section). A few of the benchmarks such as the strategic
company program contain disjunction in the head; to these our theory
does not apply. Other benchmarks were properly splittable and could
easily be translated into \NewASP following almost literally the
embedding of Theorem~\ref{theoSPLIT}. In most cases, we could split
and apply the trivial syntactic transformations exemplified in
transforming~\eqref{eq:hc-asp} to~\eqref{eq:hc-asp-fo}. Few places required  more than these trivial transformations to express \gen parts. 
Most importantly, we observed
that in \emph{all} our experiments, the \Dmodules obtained after splitting
yielded total definitions. Indeed, the rule sets belonged to one of
the aforementioned classes of total \Dmodules (non-recursive,
positive, stratified or locally stratified) and they clearly
expressed definitions of the head predicates in terms of the parameter
symbols.  Thus, $\infsemT{\II}{}{\cdot}$ provided a precise and correct
interpretation for  the  benchmark programs considered.

This observation provides experimental evidence for the claim that the
GDT para\-digm requires only total \Dmodules, and that the informal
semantics $\infsemT{\II}{}{\D}$ therefore suffices to cover GDT
practice. A similar observation was made by \citet{erdogan04}, who note
that 
\begin{quote}[w]e expect the rules in the define part of a program to not
add or remove potential solutions, but just to extend each of them by
adding the defined atoms appropriately.  
\end{quote}
This is obviously in
keeping with our restriction to total \Dmodules, which have a unique
stable model for each interpretation of their
parameters. To ensure this property, \citet{erdogan04} restrict
attention to \Dmodules without negated occurrences of defined
atoms---i.e., those that correspond to monotone inductive definitions
such as that of transitive closure. Using the results of \cite{den08}
allows us to be more general, by considering also stratified non-monotone inductive
definitions such as that of the satisfaction relation.

\textcolor{black}{Table~\ref{fig:ASP-FOinf} recaps the essence of the new theory $\infsemT{\II}$  of
informal semantics. Here, rows 9 and 10 give
the informal semantics of 
T- and G-modules,
whereas row 11 gives the informal semantics for definitional rules for D-modules. Row 12 specifies that the informal composition operator underlying D-modules is the one underlying inductive definitions. Row 13 gives the implicit composition operator of \NewASP itself (i.e., it explains what it means to gather a number of modules into a theory). As a
comparison to Table~\ref{fig:FOinf:form} shows, the D-module composition
operator (row 12) and the rule operator (row 11) are the only non-classical
elements. }

\begin{table}[t!]
\caption{The objective informal semantics for ASP-FO.\label{fig:ASP-FOinf}}
\begin{center}
\begin{tabular}{rccp{7cm}}
\hline\hline
& $\varphi$ & \phantom{aaaaa} & \hspace{1.2in}$\infsemT{\II}{}{\varphi}$\rule{0pt}{0.2cm}\\
\hline\hline
1&$f(t_1,\ldots,t_n)$ & &
\textcolor{black}{$\II(f)\langle \infsemFO{\II}{\formulas}{t_1},\ldots,\infsemFO{\II}{\formulas}{t_n}\rangle$
}
\rule{0pt}{0.2cm}\vspace{0.0cm}\\
\hline
2&$P(t_1,\ldots,t_n)$ & & \begin{minipage}{7cm}
$\II(P)\langle \infsemFO{\II}{\formulas}{t_1},\ldots,\infsemFO{\II}{\formulas}{t_n}\rangle$
\end{minipage}\rule{0pt}{0.2cm}\vspace{0.0cm}\\
\hline
3&$\varphi \lor \psi$ & & $\infsemT{\II}{}{\varphi}$ or $\infsemT{\II}{}{\psi}$ (or both)\rule{0pt}{0.2cm}\vspace{0.0cm}\\
\hline
4&$\varphi \land \psi$ & & $\infsemT{\II}{}{\varphi}$ and
$\infsemT{\II}{}{\psi}$\rule{0pt}{0.2cm}\vspace{0.0cm}\\
\hline
5&$\lnot \varphi$ & & \begin{minipage}{7cm}
\textcolor{black}{it is not the case that  $\infsemT{\II}{}{\varphi}$\\ \emph{(i.e.,
$\infsemT{\II}{}{\varphi}$ is false)}}\end{minipage}\rule{0pt}{0.2cm}\vspace{0.0cm} \\
\hline
6&$\varphi \mim\psi$ & & \begin{minipage}{7cm}
if $\infsemT{\II}{}{\varphi}$ then
$\infsemT{\II}{}{\psi}$\\
\emph{(in the sense of material implication)}\end{minipage}\rule{0pt}{0.2cm}\vspace{0.0cm}\\
\hline
7&$\exists x\ \varphi$ & & \begin{minipage}{7cm}
there exists an $x$ in the universe of discourse such that
$\infsemT{\II}{}{\varphi}$\end{minipage}\rule{0pt}{0.2cm}\vspace{0.0cm}\\
\hline
8&$\forall x\ \varphi$ & & for all $x$ in the universe of discourse, $\infsemT{\II}{}{\varphi}$
\rule{0pt}{0.2cm}\vspace{0.0cm} \\
\hline
9&T-module $\{\varphi\}$ & & $\infsemFO{\II}{}{\varphi}$ \rule{0pt}{0.2cm}\vspace{0.0cm} \\
\hline
10&G-module $\mathcal{G}$ & & $\infsemFO{\II}{}{Gcompl(\mathcal{G})}$ \rule{0pt}{0.2cm}\vspace{0.0cm} \\
\hline
11&$A \leftarrow \varphi$ & & \begin{minipage}{7cm}
if $\infsemT{\II}{}{\varphi}$ then
$\infsemT{\II}{}{A}$\\
\emph{(in the sense of definitional implication)}\end{minipage}\rule{0pt}{0.2cm}\vspace{0.0cm}\\
\hline
12&\begin{minipage}{3cm}\begin{center}
D-module\\
$\D=\{r_1,\ldots,r_n\}$
\end{center}
\end{minipage} & & \begin{minipage}{6.5cm}
\rule{0pt}{0.2cm} The relations $\II(Def(\D))$ are defined in terms of $\II(Par(\D))$ by the
 following (simultaneous) induction:
\begin{itemize}
\setlength{\itemsep}{0pt}
\setlength{\topsep}{0pt} 
\setlength{\parsep}{0pt} 
\setlength{\parskip}{0pt}
\setlength{\partopsep}{0pt}
 \item $\infsemT{\II}{}{r_1}$
 \item \ldots
 \item $\infsemT{\II}{}{r_n}$
 \end{itemize}
\vspace{0pt}
 \end{minipage}\\
\hline
13 & \begin{minipage}{3cm}
\begin{center}
\NewASP theory\\
$T = \{M_1,\ldots,M_n\}$
\end{center}\end{minipage}\rule{0pt}{0.2cm}& & $\infsemT{\II}{}{M_1}$ and 
\ldots\ and $\infsemT{\II}{}{M_1}$\vspace{0.0cm} \\
\hline
 \end{tabular}
\end{center}
\end{table}

\section{\NewASP as a classical logic}
\label{SecASPFOClassical}

The presented informal semantics $\infsemT{\II}$ is for the most part classical. Thus, we expect \NewASP to  share many properties with  FO. If not, then  $\infsemT{\II}$ should be easily refutable. In this section, we investigate a number of FO properties in the context of \NewASP.
\ignore{
The presented informal semantics $\infsemT{\II}$ is for the most part classical. Thus, it seems reasonable to expect \NewASP to also be for the most part a classical logic, in the sense that its formal semantics should share many of the properties of FO's formal semantics. Indeed, if this were not the case, then $\infsemT{\II}$ should be easily refutable. In this section, we illustrate  that \NewASP  shares a number of FO's formal properties. }

The following simple property is a direct consequence
of our definitions.

\begin{proposition}
\label{prop:minor}
Let $\Phi$ be an \NewASP module over vocabulary $\Sigma$.
If $\I, \J$ are two interpretations such that 
$\I|_{\Sigma}=\J|_{\Sigma}$, then $\I \models \Phi$ if and only if $\J\models 
\Phi$.
\end{proposition}

%
This result  states that, like an FO formula, an ASP-FO module does not impose 
constraints on symbols that do not appear in it, i.e., any expansion of a model of an ASP-FO module is again a model. In particular, \NewASP has no implicit global closed world assumption (CWA). 

A key property of FO is that substituting a formula $\varphi$ for a formula~$\psi$ that is equivalent to $\varphi$ preserves equivalence. We should hope that the same proposition holds in ASP-FO.  The following theorem states this property for  \Tmodules and \Gmodules.

\ignore{Equivalence was introduced in the new version of the paper. 
\begin{definition}
\FO formulas $\psi, \varphi$ are \emph{equivalent}, denoted by $\psi \equiv \varphi$, if for every 
structure $\I$, $\psi^{\I}=\varphi^{\I}$. 
\end{definition}
}

\begin{theorem}[Substitution property for G- and T-modules]
Let $\psi, \varphi$ be two equivalent \FO formulas. Let $T'$ be 
obtained from an \NewASP theory $T$ by substituting any number of occurrences of $\psi$ 
for $\varphi$ in \Tmodules and in the bodies of rules in \Gmodules. 
Then $T$ and $T'$ have the same models.  \label{theo:substit-GT}
\end{theorem}
Proof. This is a consequence of the substitution property in \FO and the fact 
that \Tmodules are \FO formulas and \Gmodules are equivalent to \FO formulas 
through the transformation of Corollary~\ref{cor:gmodulesFO}. 
\hfill QED

\ignore{
\begin{definition}
\FO formulas $\psi(\x), \varphi(\x)$ are \emph{equivalent}, denoted by $\psi(\x) \equiv \varphi(\x)$, if for every 
structure $\I$ and every \textcolor{red}{variable assignment} $\theta$, $\psi(\x)^{(\I,\theta)}=\varphi(\x)^{(\I,\theta)}$. 
\end{definition}

\begin{theorem}[Substitution property for G- and T-modules]
Let $\psi(\x), \varphi(\x)$ be two equivalent \FO formulas. Let $T'$ be 
obtained from an \NewASP theory $T$ by substituting any number of occurrences of $\psi(\x)$ 
for $\varphi(\x)$ in \Tmodules and in the bodies of rules in \Gmodules. 
Then $T$ and $T'$ have the same models.  \label{theo:substit-GT}
\end{theorem}
Proof. This is a consequence of the substitution property in \FO and the fact 
that \Tmodules are \FO formulas and \Gmodules are equivalent to \FO formulas 
through the transformation of Corollary~\ref{cor:gmodulesFO}. 
\hfill QED
}

\smallskip The situation is less straightforward for \Dmodules, where
the formal semantics depends on the concept of the Kleene three-valued
truth assignment \citep{Kleene52}. 
As explained in the discussion following Definition~\ref{defdef},
 Kleene's three-valued truth assignment can be derived from the
notion of the satisfaction relation for pairs of interpretations. 
Nevertheless, here it is useful to give the explicit definition. 

Let $\tilde{\I}$ be a \emph{three-valued}
structure, i.e., one which interprets each predicate symbol $P/n$ as a
function from $dom(\tilde{\I})^n$ to the set of truth values
$\{\Tr,\Fa,\Un\}$.  We order these truth values under the truth order as $\Fa \leqt \Un \leqt
\Tr$ and define the complement operator $\Fa^{-1} = \Tr$, $\Tr^{-1}
=\Fa$ and $\Un^{-1} = \Un$.  Proceeding by induction in a similar way
as in Definition \ref{deftrue}, we define the truth value
$\vph^{\tilde{\I},\theta}$ of a formula $\vph$ with respect to
$\tilde{\I}$. This definition
follows Kleene's weak truth tables.  \newcommand{\kl}[2]{{#1}^{{#2}}}
\begin{itemize}
\item[$-$] $\kl{P(t_1,\dots,t_n)}{\tilde{\I}}  :=  P^{{\tilde{\I}}}(t_1^{{\tilde{\I}}},\dots,t_n^{{\tilde{\I}}})$;
\item[$-$] $\kl{(\neg \psi)}{{\tilde{\I}}} := (\kl{\psi}{{\tilde{\I}}})^{-1}$;
\item[$-$] $\kl{(\psi\land\varphi)}{{\tilde{\I}}} := Min_\leq(\kl{\psi}{{\tilde{\I}}},\kl{\varphi}{{\tilde{\I},}})$;
\item[$-$] $\kl{(\psi\lor\varphi)}{\tilde{\I}} := Max_\leq(\kl{\psi}{{\tilde{\I}}},\kl{\varphi}{{\tilde{\I}}})$;
\item[$-$] $\kl{(\exists x\ \psi)}{\tilde{\I}} := Max_\leq( \{ \kl{\psi}{{\tilde{\I}[x:d]}} | d \in D\})$;
\item[$-$]  $\kl{(\forall x\ \psi)}{{\tilde{\I}}} := Min_\leq( \{ \kl{\psi}{{\tilde{\I}[x:d]}} | d \in D\})$.
\end{itemize}
To link this definition with Definition~\ref{defdef}, each three-valued 
structure $\tilde{\I}$ corresponds to a pair $(\I_l,\I_u)$ of two-valued 
structures. To obtain $\I_l$ and $\I_u$ from $\tilde{\I}$,  $\Un$ is mapped 
to $\Fa$ and to $\Tr$, respectively. Consequently, $\I_l$ represents a lower 
approximation of the (two-valued) structures represented by $\I$, and 
$\I_u$ represents an upper approximation. The relationship between 
$\varphi^{\tilde{\I}}$ and $(\varphi^{(\I_l,\I_u)}, \varphi^{(\I_u,\I_l)})$ is given
by the following correspondences: $\Tr \leftrightarrow (\Tr,\Tr)$, 
$\Fa \leftrightarrow (\Fa, \Fa)$, and $\Un \leftrightarrow (\Fa,\Tr)$. 
The tuple $(\Tr,\Fa)$ does not arise since $\I_l\leqt\I_u$.

\begin{definition}
We call \FO formulas $\psi, \varphi$ \emph{3-equivalent}, denoted  $\psi \equiv_3 \varphi$, if for every three-valued structure $\tilde{\I}$ interpreting all symbols of $\psi, \varphi$, 
$\psi^{\tilde{\I}}=\varphi^{\tilde{\I}}$. 
\end{definition} 

Two 3-equivalent \FO formulas are also equivalent since (two-valued)
interpretations are a special case of three-valued interpretations. The
inverse is not true and some properties of \FO, such as the law
of excluded middle, do not hold in three-valued logic.  
For instance,
$\top$ (true) and $\varphi\lor\neg\varphi$, or $\varphi$ and
$(\varphi\land\psi)\lor(\varphi\land\neg\psi)$ are not
3-equivalent. However, most standard equivalences are also
3-equivalences:
\begin{itemize}
\item[$-$] $\neg\neg \varphi \equiv_3 \varphi$ (double negation);
\item[$-$] $\neg(\varphi\land\psi) \equiv_3 \neg\varphi\lor\neg\psi$ 
(De Morgan);
\item[$-$] $\neg(\varphi\lor\psi) \equiv_3 \neg\varphi\land\neg\psi$ 
(De Morgan);
\item[$-$] $(\varphi\land\psi)\lor (\neg\varphi\land\neg\psi)\equiv_3 (\neg\varphi\lor\psi)\land (\varphi\lor\neg\psi)$ (these are two rewritings of 
$\varphi\Leftrightarrow\psi$);
\item[$-$] $\neg\forall \x\ \varphi \equiv_3 \exists \x\ \neg\varphi$;
\item[$-$] $\neg\exists \x\ \varphi \equiv_3 \forall \x\ \neg\varphi$;
\item[$-$]  distributivity laws, commutativity and associativity laws, idempotence, etc.
\end{itemize}
\ignore{
All of these have a straightforward proof. The propositional ones can be  checked by the method of (3-valued) truth tables.  The two remaining ones 
follow directly from the property that for two formulas $\varphi, 
\psi$ to be 3-equivalent it suffices that that $\varphi', \psi'$ are 
equivalent, where $\varphi', \psi'$ are obtained by substituting new 
predicate symbols $P'$ for symbols $P$ in all negative occurrences of 
$P$. 
}

\begin{theorem}[Substitution property for D-modules] \label{theosub1}
Let formulas $\psi$ and $\vph$ be 3-equivalent. If an \NewASP 
theory $T'$ is obtained from an \NewASP 
theory $T$ by substituting any number of occurrences of $\psi$ for $\varphi$ 
in bodies of rules in \Dmodules, then $T$ and $T'$ have the same models. 
\end{theorem}
Proof. In \citep{pelov}, it was shown that the parametrized stable models of a \Dmodule $\D$ can be characterized as  a specific kind of  fixpoints, called stable fixpoints, of the three-valued immediate consequence operator associated to $\D$. Since any substitution of a formula by a 3-equivalent formula preserves the operator, it also preserves its stable models. Consequently, models are
preserved, too. \hfill QED

\smallskip Thus, most standard \FO transformations are equivalence
preserving D-modules as well.  In other words, virtually all standard
``laws of thought'' apply in \NewASP: the De Morgan laws, double
negation, distributivity, associativity, commutativity, idempotence.
This property of \NewASP has deep practical implications. It means
that the programmer has (almost) the same freedom as in \FO to express
an informal proposition in \NewASP. It also implies that the
correctness of the programmer's formalization does not depend on
subtleties of the formalization that go beyond common understanding.

Here is an example of a standard \FO transformation that is not
equivalence preserving in the context of a D-module. 
Since the law of excluded middle does not hold in 3-valued logic, the
formulas $\top$ and $P\lor \neg P$ are equivalent but not
3-equivalent. Substituting the second for the first in the body of the
rule of the \Dmodule:
\[ \{ P\rul \top\}\]
 results in the non-equivalent \Dmodule:
\[ \{ P\rul P\lor\neg P\}.\] 

Indeed, the first module has a unique model $\{P\}$, while the second
has no models. Thus, reasoning by cases does not in general preserve
equivalence in \Dmodules. 
%
%
The explanation of this phenomenon lies in the nature of inductive
definitions 
{(and not in, e.g., the nature of negation in \Dmodules). Indeed,
(inductive) definitions are sensitive to negated propositions in
rule bodies, because such propositions constrain the order in which rules may
be applied \citep{KR/DeneckerV14}.} Therefore, rewriting rule bodies while adding such
propositions may disturb the rule application process in an
irrecoverable way and turn a sensible definition in a non-sensible
one. \citet{KR/DeneckerV14} demonstrate that similar phenomenon can be observed in mathematical texts.

While this example shows that reasoning by cases cannot be applied in
general, the following theorem illustrates that any equivalence preserving
transformation of classical logic, including reasoning by cases, can
be applied to \Dmodule rule bodies, provided it is used  with care. In
particular, the transformation should not destroy the totality of the
definition.
\begin{theorem}[Second substitution property for \Dmodules] \label{theosub2}
Let $\psi\equiv\varphi$ and let~$T'$ be obtained from the \NewASP 
theory~$T$ by substituting occurrences of~$\psi$ for~$\varphi$ 
in the bodies of rules in \Dmodules. If $T$ and $T'$ are both total, then
$T$ and $T'$ have the same models.
\end{theorem}
Proof. Also this theorem was proven by \citet{pelov}. In fact, it is a consequence of a 
more general property \citep{pelov} that if two \Dmodules have the same 2-valued immediate consequence operator, 
then the well-founded models of both may be different but they are not 
contradicting each other. That is: there are no atoms that are true in one and false in the other. Any application of an equivalence preserving rule on  a body of a \Dmodule  obviously preserves the 2-valued immediate consequence operator.  If both \Dmodules are total, their well-founded models are 2-valued and hence, identical. These models are also the unique stable models of the two \Dmodules. 
\hfill QED

These results essentially show that we are free to apply any equivalence preserving transformation of FO to the rule bodies of a D-module, as long as we are careful not to turn the D-module into a nonsensical inductive definition.

\section{Related Work and Discussion}
\label{SecDiscussion}

This section discusses the scope of the results in this article and situates them within the ASP literature. 

\paragraph{Extending the \coreASP language}
A limitation
of the \coreASP language studied in this article is that it lacks  aggregates or weight
constraints. Indeed, such constructs are used in many ASP
applications. \citet{pelov} extend FO(ID) 
with aggregates. That work can be adopted ``verbatim'' to the case of the
logic \NewASP. \textcolor{black}{Importantly, extending $\infsemT{\II}{}{\varphi}$ to
theories with aggregates is also not problematic.} A clear-cut
example of a \Dmodule involving induction over aggregates is the
following definition specifying that a company $x$ controls a company
$y$ if the sum of the shares of $y$ that $x$ owns directly and of the
shares of $y$ owned by companies $c$ controlled by $x$ is more than
50\%.
\[ \defin{ \forall x (\Cont(x,y) \rul 50 < \\ \ \ \ \ \ Sum\left\{(s,c) : 
\begin{array}{l}
(c=x \land Shares(x,y,s)) \lor\\
(\Cont(x,c) \land Shares(c,y,s))
\end{array}
\right\})}\]
This is an example of a monotone inductive definition with recursion
over aggregate expressions.  Under the intended informal semantics for
the $Sum$ aggregate, 
$\infsemT{\II}{}{\cdot}$ produces the following informal reading of this
\Dmodule:
\begin{quote}
The relation ``$x$ controls $y$'' is defined in terms of the relation ``$x$
holds $s$ percent of shares in $y$'' by the following
induction:
\begin{itemize}
\item \textcolor{black}{Consider the sum  of the percentages $s$ for companies $c$ such that either $c$ and $x$ are the same and $x$
  holds $s$
  percent of shares in $y$ or $x$ controls $c$ and $c$  holds $s$
  percent of shares in $y$.  If this sum is greater then 50, then $x$ controls~$y$.}
\end{itemize}
\end{quote}
This informal inductive definition provides a precise and meaningful
interpretation of the recursion over an aggregate in the 
\Dmodule above.

\paragraph{Links to other developments in ASP}


\citet{pea97} proposed to use the logic of Here and There (HT)
as a meta-logic to study ASP semantics. Pearce's work maps an ASP
program to a theory in HT and characterizes its answer sets as a
specific subclass of the models of this theory, called
equilibrium models. \citet{pea04,pea05} generalized
these ideas to arbitrary first order formulas.  
Also \citet{fer09} conservatively lifted ASP to the full FO syntax using
a form of circumscription -- an operator {\sc sm} defined in second
order logic.  These characterizations proved to be useful for
analyzing and extending ASP semantics.  For example, the logic HT was
shown to underlie the notion of strong equivalence of programs
\citep{lif01,lif07a}, while the use of operator {\sc sm} provided an
elegant alternative to the definition of stable models of logic
programs containing choice rules. The relation between these different approaches was
investigated in \cite{lin11}. 

While these characterizations are powerful formal tools (e.g., our
Theorem \ref{theoSPLIT} follows from a result proved by \cite{fer09}
about the {\sc sm} operator), they do not in themselves directly contribute to the understanding of the
informal semantics of ASP. For example, neither the informal semantics of
HT nor of equilibrium logic has been developed so far. Similar arguments apply to the
semantics of logic programs under operator {\sc sm}, where the effect
of this operator on the informal semantics of the formulas has not yet
been studied.

 Several features of \NewASP also appear in other variants
 of ASP. As we observed earlier, the intuitive structure of a
 GDT-program is \cred{hidden}\ignore{flattened out} in an ASP program. 
Techniques developed in ASP to cope with this include splitting to detect
 natural components of the program
 \citep{litu94,JanhunenOTW09,fer09a}, and module systems, e.g., in
 \citep{gel02,oik08,lt13}.  

Due to its non-Herbrand semantics,
 \NewASP  imposes neither  Domain Closure Axiom nor the Unique Names
 Axiom. Thus, function symbols that are not constrained by a Herbrand
 module act as  function symbols in classical logic. Recent extensions
 of ASP  with similar features are open domain ASP logics such as
 those of \citet{fer09,LifschitzPY12} and ASP with functions
 \citep{LinW08,Balduccini12,Cabalar11}.  The progression semantics of
 \citet{zhang10} and \citet{zhou11} also allows non-Herbrand models and
 makes a distinction between the intensional and extensional
 predicates of a theory. The ordered completion semantics 
 \citep{asuncion12} provides another way \cred{to define}\ignore{of defined} ASP for
 non-Herbrand models, which has also been extended to aggregates \citep{asuncion12}.  A detailed comparison with these languages would be interesting but is beyond the scope of this paper.

\paragraph{Well-founded versus stable semantics} Comparisons between
the well-founded and stable semantics have long been the subject of
discussion. Our position is that once
these semantics are generalized to their parametrized versions and the
internal structure of a program (in particular its \define components) is identified, 
then the differences between both semantics disappear for practically
relevant programs.  They are just different mathematical
formalizations for the same informal semantics: sets of clauses that
{\em define} certain predicates/relations in terms of parameter symbols.  However,
while the two semantics are equivalent in the case of sensible 
(i.e., total) definitions, they are not equivalent in the case of other rule
sets. 
The well-founded semantics identifies non-total definitions by producing a 3-valued model.  In the stable semantics, nonsensical 
definitions are revealed by the absence of stable models or by multiple
stable models. However, some programs have a unique stable model, but are
not sensible definitions. For instance, the following logic program with three defined
symbols $a$, $b$ and $f$ and no parameter symbols
\[ \defin{
f \rul \neg f \land b\\
a \rul \neg b\\
b \rul \neg a
}\]
has a unique stable model $\{a\}$ but does not represent a sensible 
definition of  $a$, $b$, and~$f$. 

In practice, if ASP programmers avoid rule sets with cycles over
negation, definitions are total and the two semantics coincide. The
evolving GDT-programming methodology in ASP discourages cycles over 
negation as an encoding technique in favour of more direct representations 
based on the use of choice rules. This means that the debate between
both semantics is losing its practical relevance.

\paragraph{Tools for \NewASP and FO(ID)}

\NewASP is more than a theoretical mechanism for a semantic study of
GDT programs. It is also a viable logic for which efficient tools
already exist. 

Similarly to FO and FO(ID), \NewASP is an open domain logic and its
models can be infinite. In general, its satisfiability problem is
undecidable (and not just co-semidecidable) --- this can be
proved by adapting the corresponding result concerning the logic
FO(ID) \citep{den08}. In many search problems, however, a finite domain
is given. That opens a way to practical problem solving. One can
apply, e.g., finite model checking, finite satisfiability checking,
finite Herbrand model generation or, in case the domain and data are
available as an input structure, {\em model expansion} \citep{MitchellT05}.

Answer set programming solvers such as {\rm
  Smodels}~\citep{nie00}, DLV~\citep{dlvtocl06}, {\sc cmodels}~\citep{giu06},
 {\em clasp}~\citep{geb07}, and WASP~\citep{AlvianoDFLR13} can be viewed as
systems
supporting a subset of \NewASP (modulo the rewriting that we
proposed). Also, \NewASP/FO(ID) is formally an extension of Abductive Logic
Programming, and hence, abductive reasoners such as the solver
$\mathcal{A}$-system \citep{ijcai/KakasND01} can be seen to implement abductive
reasoning for a fragment of \NewASP/FO(ID). 

Several systems have
been designed with the intention to support extensions or variants of
FO with rules.  An early finite Herbrand model generator that
supported a fragment of \NewASP/FO(ID) was developed by East and
Truszczynski (\citeyear{tocl/EastT06}). This system supports clausal
logic and a set of definite Horn rules under the minimal model
semantics; in our terminology this is a negation-free \Dmodule
representing a monotone inductive definition. Another solver is the
Enfragmo system \citep{AavaniWTTM12} that supports model expansion for
FO and non-recursive definitions. Both systems support
aggregates.
Also,
Microsoft's system FORMULA \citep{conf/ictac/JacksonS13} supports a
form of satisfiability checking for a similar logic.

At present, the IDP system \citep{IDP,IDP2} offers the most complete
implementation of \NewASP as well as FO(ID).  IDP is a knowledge base
system that provides various forms of inference, including model
expansion and Herbrand model generation.  From a KR point of view, the
system was developed to support essentially the GDT methodology: the
representation of assertional knowledge (corresponding to G- and
T-modules) and definitional knowledge (\Dmodules). The language supported by IDP
includes FO, \Dmodules, aggregates, quantified
existential quantifiers, multiple definitions, bounded forms of
arithmetic, uninterpreted functions and constructor functions,
etc. \Gmodules are to be ``emulated'' by FO formulas. A flag can be
used to select the parametrized stable or well-founded semantics for
rule sets; this switches the system effectively between (extensions
of) \NewASP and FO(ID).

\paragraph{\NewASP and FO(ID) in a historical perspective}


Originally, logic programming was seen as the Horn fragment of
classical logic. This view soon became untenable due to negation
as failure. One of the earliest proposed explanations was the
view of logic programs as definitions. It underlied the work of
Clark (\citeyear{Clark78}) and \citeauthor{ChandraH82}
(\citeyear{ChandraH82}). It was also present in Kowalski's book
(\citeyear{Kowalski79a}).  In \citeyear{ge1}, Gelfond and Lifschitz proposed the
autoepistemic view of logic programs as a synthesis of logic
programming and nonmonotonic reasoning. This proposal
led to the development of ELP in \citeyear{ge2}.  
Note that nonmonotonic reasoning, one of the roots of ELP, had been developed fifteen
years earlier as a reaction against the shortcomings of classical logic
for common-sense knowledge representation.

The research direction set out by
\citeauthor{ChandraH82} was followed up by Apt et al.
(\citeyear{minker88/AptBW88}) and Van Gelder et
al. (\citeyear{vrs91}).  Although the link between logic programs and
inductive definitions was at the heart of these developments, it was not
 made explicit. The link was strengthened again
by \citet{jcss/Schlipf95} and later fully explicated by \citet{tocl/DeneckerBM01}; it led to the logic FO(ID)
\citep{den00,den08}. Interestingly, the latter logic grew out of a
semantic study of another knowledge representation extension of
logic programming: Abductive Logic Programming \citep{KakasKT92}. In
\NewASP and FO(ID), the main relict of logic programs is the \Dmodule
which is viewed as a (possibly inductive) definition.  Our Tarskian
perspective is a proposal to ``backtrack'' to the  early view of
logic programs as definitions. Thus, the present paper is a confluence
of many research directions. ASP arose at the meeting point of two logic
paradigms---nonmonotonic reasoning and logic programming---that in origin
were antitheses to classical logic. One of the contributions here is
to reconcile ASP with the objective informal semantics of classical
logic.  This is the crucial step towards a synthesis of these
languages, as is achieved in \NewASP and FO(ID).

\section{Conclusion}
\label{sec:concl}






The goal of this paper was to develop a theory of the \emph{informal
  semantics of ASP} that: 
\begin{itemize}
	\item 
 explains the ASP practice of
GDT programming, and
\item  matches  the informal reading of an ASP
expression with  the informal proposition that the human programmer
has in mind when he writes it. 
\end{itemize}
To conduct our analysis, we presented
the formalism \NewASP, whose modular structure is geared specifically
towards the GDT paradigm and in which the internal structure of GDT
programs is made explicit. By reinterpreting answer sets as objective 
possible worlds rather than as belief sets, we obtained an informal
semantics for the GDT fragment of ASP that combines modules by means
of the standard conjunction, and captures the roles of different
modules in GDT-based programs. This allowed us to clarify the nature of the three
sorts of conditionals found in ASP, and of the negation symbol in G,
D, and T-modules. In addition, the close connection between \NewASP
and FO(ID) assisted us in providing, to the best of our knowledge, the
first argument for the correctness of the stable model semantics as a
formalization of the concept of an (inductive) definition.

A study of a logic's informal semantics is an investigation into the foundations of the logic. We showed that explaining ASP from a Tarskian point of view has a deep impact on our view of the ASP language. All together, our study forces us to reconsider the intuitive meaning of ASP's basic connectives, it  redefines ASP's position in the spectrum of logics and it shows much tighter connections with existing logics including FO and FO(ID).

\section*{Acknowledgments}
The first and fourth author are supported by Research
Foundation-Flanders (FWO-Vlaanderen) and by GOA 2003/08 "Inductive
Knowledge Bases". The second author was supported by a CRA/NSF 2010
Computing Innovation Fellowship and FRI-2013: Faculty Research
International Grant.  The third author was supported by the NSF grant
IIS-0913459. For sharing their views on topics related to this paper,
we would like to thank the following people: Maurice Bruynooghe,
Vladimir Lifschitz, Michael Gelfond and Eugenia Ternovska.


\bibliographystyle{elsarticle-harv}
\bibliography{asp-fo}

\section*{Appendix}

To prove Theorem~\ref{theoSPLIT}, we rely on results for the extension
of ASP to the syntax of classical logic and to arbitrary
interpretations presented in \cite{fer09}. For
an \FO sentence $\Pi$ over a finite vocabulary $\Voc$ and a finite set
$\bold{p}\subseteq\Voc$ of predicate symbols, \citeauthor{fer09}
introduced a second-order formula $SM_\bold{p}(\Pi)$ and defined a (possibly non-Herbrand) structure $\M$ to be a {\em general answer set} of $\Pi$ relative to
$\bold{p}$ if $\M$ is a model of $SM_\bold{p}(\Pi)$. They defined a
general answer set of $\Pi$ as a general answer set of~$\Pi$ relative
to the set of all predicates in $\Voc$. We refer to the paper by
\citeauthor{fer09} for details; the actual definition of the operator
$SM_\bold{p}$ is not important for our argument and so we omit it.

\newcommand{\fer}[1]{\tilde{#1}}

The \coreASP language is embedded in this generalized formalism by a
modular transformation. The transformation maps constraints $\rul
L_1,\dots,L_n$ to the same \FO sentences as in \NewASP: $\neg \exists
\xxx (L_1\land\dots\land L_n)$. It maps rules $p(\ttt)\rul
L_1,\dots,L_n$ to formulas $\forall\xxx(L_1\land\dots\land
L_n\Rightarrow p(\ttt))$ and it maps choice rules $\{p(\ttt)\}\rul
L_1,\dots,L_n$ to formulas $\forall\xxx(\neg\neg p(\ttt)\land
L_1\land\dots\land L_n\Rightarrow p(\ttt))$. The mapping of a rule $r$
is denoted $\fer{r}$. The embedding of any set $\Pi$ of rules is the
conjunction of the mapping of its rules and is denoted $\fer{\Pi}$.

\citeauthor{fer09}~(\citeyear{fer09}) defined a structure $\M$ to be 
a general answer set of a \coreASP program $\Pi$ if $\M$ is a general 
answer set of the FO sentence $\fer{\Pi}$ (that is, if $\M$ satisfies 
$SM_{\Voc_P}(\fer{\Pi})$). 
An 
{\em answer set} of a \coreASP program $\Pi$  is a 
Herbrand structure $\M$ that is a general answer set of $\Pi$ (an Herbrand 
structure that satisfies $SM_{\Voc_P}(\fer{\Pi})$).

 
\begin{theorem} \label{theoSPLIT-a}
  Let $\Pi$ be a \coreASP program over a finite vocabulary $\Voc$ 
  and $\{\Pi_1,\dots,\Pi_n\}$ be a proper splitting of $\Pi$. Then an 
  interpretation $\M$ is a general answer set of $\Pi$ if and only if
  $\M$ is a model of the \NewASP theory $\{\wh{\Pi}_1,\dots,\wh{\Pi}_n,(Def,\{\})\}$. 
\end{theorem}
 Theorem~\ref{theoSPLIT-a} is a  generalization of Theorem~\ref{theoSPLIT} since it holds  for (non-Herbrand) general answer sets. 

\smallskip
\noindent
Proof. 
We will apply the Symmetric Splitting Theorem \cite{fer09a}. That 
theorem is stated in the language of arbitrary FO sentences. It applies
to finite programs under the rewriting of rules as sentences we 
discussed above.

By definition, a structure $\M$ is a general answer set of
$\Pi$ if and only if it 
satisfies
$SM_{\Voc_P}(\fer{\Pi})$  or, equivalently, of
$SM_{\Voc_P}(\fer{\Pi}_1\land\cdots\land\fer{\Pi}_n)$.

Without loss of generality, we assume that in the splitting $\{\Pi_1,
\ldots,\Pi_n\}$, programs $\Pi_1,\ldots,\Pi_i$ consist of choice
rules, $\Pi_{i+1},\ldots,\Pi_j$ of normal program rules, and
$\Pi_{j+1},\ldots, \Pi_n$ of constraints. Since the sentences
corresponding to constraints are of the form $\neg \psi$, results of
\citeauthor{fer09}~(\citeyear{fer09}) imply that
$SM_{\Voc_P}(\fer{\Pi})$ is equivalent to
\[
SM_{\Voc_P}(\fer{\Pi}_1\land \cdots\land \fer{\Pi}_j)\land\fer{\Pi}_{j+1}\land\cdots
\land\fer{\Pi}_{n}.
\]

Next, we observe that $\fer{\Pi}_1\land\cdots\land\fer{\Pi}_j$ can be 
written as $\fer{\Pi}_1\land\cdots\land\fer{\Pi}_j\land \top$. Moreover, 
since $\{\Pi_1,\dots,\Pi_n\}$ is a proper splitting of $\Pi$, the sentences
$\fer{\Pi}_1,\cdots,\fer{\Pi}_j,\top$ together with the sets $\hd(\Pi_1),
\ldots, \hd(\Pi_j),\Voc_P\setminus\hd(\Pi)$ of predicates satisfy the
assumptions of the Symmetric Splitting Theorem \cite{fer09a}.
Consequently, $SM_{\Voc_P}(\fer{\Pi}_1\land\cdots\land\fer{\Pi}_j)$ is 
equivalent to
\[
SM_{\hd(\Pi_1)}(\fer{\Pi}_1)\wedge\cdots\wedge SM_{\hd(\Pi_j)}(\fer{\Pi}_j)
\land SM_{\Voc_P\setminus\hd(\Pi)}(\top).
\]

To deal with choice rules in a program $\Pi_k$, $1\leq k\leq i$, we
use the generalized Clark's completion \cite{fer09} defined for the
case of FO sentences. To describe it, recall that $\fer{\Pi}_k$
consists of sentences of the form $\forall \xxx\ (\neg \neg p(\ttt)
\land \vph \mim p(\ttt))$.
Each such sentence is first rewritten as $\forall\yyy\ (\neg \neg
p(\yyy)\land \exists\xxx\ (\ttt=\yyy \land \vph) \mim p(\yyy))$, where
$\yyy$ are fresh variables.  Next, we combine all formulas obtained in
this way into a single one, which has the form
\[
\forall \yyy\ (\neg\neg p(\yyy) \land (\psi_1\lor\cdots\lor
\psi_s)\mim p(\yyy)),
\]
where $s$ is the number of sentences in
$\Pi_k$. The generalized Clark's completion of the original set $\Pi_k$ is obtained from this sentence by substituting equivalence for implication:
\[
\forall \yyy\ (\neg\neg p(\yyy) \land (\psi_1\lor\cdots\lor \psi_s)
\Leftrightarrow p(\yyy)).
\]
One can  verify that this sentence is equivalent in \FO to the sentence 
\[
\forall\yyy\ (p(\yyy) \mim (\psi_1\lor\cdots\lor\psi_s)).
\]
Since the splitting $\{\Pi_1,\ldots,\Pi_n\}$ is proper, all rules of $\Pi_k$
have the same predicate in their heads, and this predicate has no ocurrences
in the bodies of rules of $\Pi_k$. Consequently, $\Pi_k$ is \emph{tight} 
(for a definition of tightness we refer to \citet{fer09}). 
By the result on tight programs proved by \citet{fer09}, 
models of $SM_{\hd(\Pi_k)}(\fer{\Pi}_k)$ and models of the Clark's completion 
of $\fer{\Pi}_k$ coincide.

We now note that the process of completion applied we described in the 
section on the semantics of the logic \NewASP, when applied to the 
\Gmodule $\Pi_k$ (that is, formally, the \Gmodule $(\hd(\Pi_k),\Pi_k)$), 
results in an equivalent G-module $\{\forall \yyy\ p(\yyy)\leftarrow 
(\psi_1\lor\cdots \lor \psi_s)\}$
(cf. Theorem \ref{prop:ch-ra} and the discussion that precedes it). By 
Theorem \ref{prop:ch-r}, models of that G-module coincide with models of the
sentence $\forall\yyy(p(\yyy) \mim (\psi_1\lor\cdots\psi_s))$. Thus,
models of $SM_{\hd(\Pi_k)}(\fer{\Pi}_k)$ and of the G-module $\Pi_k$ are
the same.

Next, we observe that programs $\Pi_k$, $i+1\leq k\leq j$, are normal. 
As a consequence of the results by \citeauthor{Truszczynski12}
(\citeyear{Truszczynski12}), we obtain that models of 
$SM_{\hd(\Pi_k)}(\fer{\Pi}_k)$ and of the \NewASP D-module $\fer{\Pi}_k$ 
coincide. 

Finally, models of $SM_{\Voc_P\setminus\hd(\Pi)}(\top)$ are precisely
those interpretations of $\Voc$ that interpret each predicate symbol
in $\Voc_P\setminus\hd(\Pi)$ with the empty relation. 
Applying the generalized Clark's completion to $\top$ (with respect
to the vocabulary $\Sigma_P\setminus\hd(\Pi)$) results in the FO 
sentence
\beq
\bigwedge_{Q\in \Voc_P\setminus\hd(\Pi) }\forall \yyy\ (Q(\yyy)\Leftrightarrow\bot)
\eeq{eq:compyyy}
Note that $\top$ is a tight sentence~\cite{fer09} and hence models of
$SM_{\Voc_P\setminus\hd(\Pi)}(\top)$ and~\eqref{eq:compyyy} coincide.
It is easy to see that models of $\cwa(\Sigma_P\setminus\hd(\Pi))$ coincide with models of~\eqref{eq:compyyy}.

Gathering all earlier observations together, we obtain that models of 
$SM_\bold{p}(\fer{\Pi})$, that is answer sets of $\Pi$, coincide with 
models of the \NewASP theory $\{\Pi_1,\ldots,\Pi_n,
\cwa(\Sigma_P\setminus\hd(\Pi))\}$. \hfill QED

\smallskip
 Theorem~\ref{theoSPLIT} is a corollary of the above result limited to Herbrand structures.

\smallskip
\noindent
{\bf Theorem~\ref{theoSPLIT}~}{\it
	Let $\Pi$ be a \coreASP program over a finite vocabulary $\Voc$
	with a proper splitting $\{\Pi_1,\dots,\Pi_n\}$. Then an
	interpretation $\M$ is an answer set of $\Pi$ if and only if~$\M$ is a model of the \NewASP theory $\{\Her{\Voc_F},\wh{\Pi}_1,\dots,\wh{\Pi}_n,(Def,\{\})\}$, where $Def = \Voc_P\setminus\hd(\Pi)$.
}

\smallskip
\noindent
Proof: The result follows from Theorem \ref{theoSPLIT-a} and from 
the observations that answer sets are Herbrand interpretations and
the only interpretations that satisfy the Herbrand module $\Her{\Voc_F}$ 
are Herbrand ones. \hfill QED

\end{document}


\end{document}
\grid
\grid